\documentclass{article}

    \PassOptionsToPackage{numbers, compress}{natbib}

    \usepackage[final]{neurips_2022}

\usepackage[utf8]{inputenc} %
\usepackage[T1]{fontenc}    %
\usepackage{hyperref}       %
\usepackage{url}            %
\usepackage{booktabs}       %
\usepackage{amsfonts}       %
\usepackage{nicefrac}       %
\usepackage{microtype}      %
\usepackage{notation}
\usepackage{researchpack}

\usepackage[usenames,dvipsnames]{xcolor}
\usepackage[title]{appendix}
\usepackage{caption}
\usepackage{subcaption}
\usepackage{color,soul,booktabs}
\usepackage{stmaryrd}
\usepackage{tikz}
\usepackage{xspace}
\usepackage{graphicx}
\usepackage{amsmath}
\usepackage{amssymb}
\usepackage{mathtools}
\usepackage{dsfont}
\usepackage[linesnumbered,lined,boxed,commentsnumbered,ruled,vlined, noend]{algorithm2e}
\usepackage{multirow}
\usepackage{bbm}
\usepackage{bm}
\usepackage{wrapfig}

\usepackage{tikz}
\usetikzlibrary{pc}
\usetikzlibrary{positioning}
\usetikzlibrary{trees}
\usetikzlibrary{arrows,shapes,shapes.geometric,backgrounds}
\usetikzlibrary{calc}
\usetikzlibrary{decorations.markings}
\usetikzlibrary{circuits.logic.US}
\usetikzlibrary{fit}
\usepackage{stmaryrd}

\DeclareMathOperator{\flow}{F}

\newcommand{\given}{\vert}

\newcommand{\prune}{\ensuremath{\mathsf{Prune}}}
\newcommand{\grow}{\ensuremath{\mathsf{Grow}}}

\newcommand{\erand}{\textsc{eRand}}
\newcommand{\eparam}{\textsc{eParam}}

\newcommand{\eflow}{\textsc{eFlow}}

\title{Sparse Probabilistic Circuits via \\Pruning and Growing}

\author{%
Meihua Dang \\
CS Department\\ UCLA\\
\texttt{mhdang@cs.ucla.edu}
\And
Anji Liu \\
CS Department\\ UCLA\\
\texttt{liuanji@cs.ucla.edu}
\And 
Guy Van den Broeck\\
CS Department\\ UCLA\\
\texttt{guyvdb@cs.ucla.edu}
}

\begin{document}

\maketitle

\begin{abstract}
Probabilistic circuits (PCs) are a tractable representation of probability distributions allowing for exact and efficient computation of likelihoods and marginals. 
There has been significant recent progress on improving the scale and expressiveness of PCs. However, PC training performance plateaus as model size increases. We discover that most capacity in existing large PC structures is wasted: fully-connected parameter layers are only sparsely used.
We propose two operations: \emph{pruning} and \emph{growing}, that exploit the sparsity of PC structures. Specifically, the pruning operation removes unimportant sub-networks of the PC for model compression and comes with theoretical guarantees. The growing operation increases model capacity by increasing the size of the latent space. 
By alternatingly applying pruning and growing, we increase the capacity that is meaningfully used, allowing us to significantly scale up PC learning.
Empirically, our learner achieves state-of-the-art likelihoods on MNIST-family image datasets and on Penn Tree Bank language data compared to other PC learners and less tractable deep generative models such as flow-based models and variational autoencoders (VAEs).
\end{abstract}

\section{Introduction}

\label{sec:intro}

\begin{wrapfigure}{r}{0.505\textwidth}
    \centering
    \includegraphics[width=0.49\textwidth]{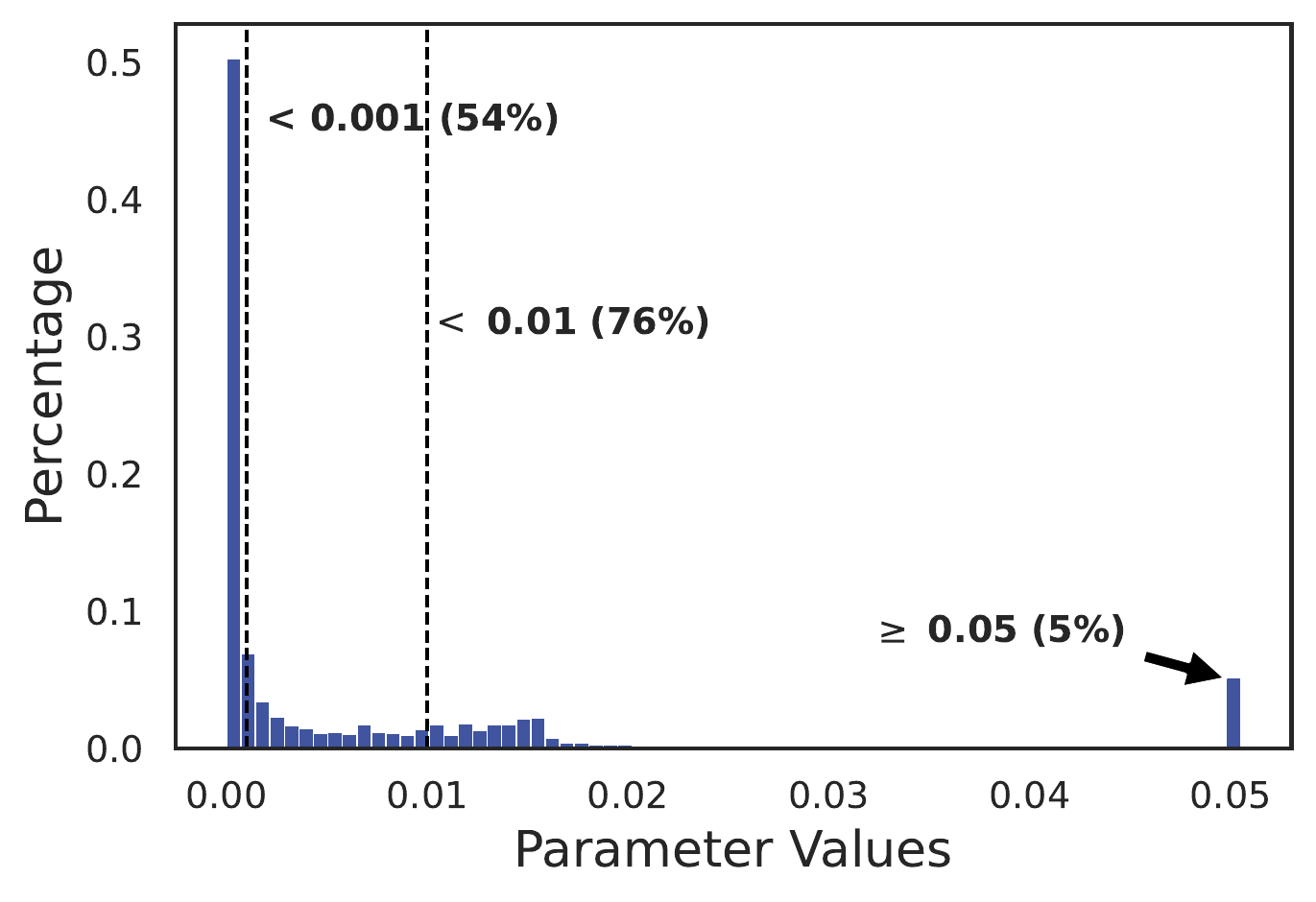}
    \caption{\label{fig:params_hist}
    Histogram of parameter values for a state-of-the-art PC with 2.18M parameters on~MNIST. 95\% of the parameters have close-to-zero values.
    }
    \vspace{-1.05em}
\end{wrapfigure}

Probabilistic circuits (PCs)~\citep{PCTuto20,ProbCirc20} are a unifying framework to abstract from a multitude of tractable probabilistic models. 
The key property that separates PCs from other deep generative models such as flow-based models~\citep{papamakarios2021normalizing} and VAEs~\citep{kingma2013auto} is their \emph{tractability}. It enables them to compute various queries, including marginal probabilities, exactly and efficiently~\citep{vergari2021compositional}. 
Therefore, PCs are increasingly used in inference-demanding applications such as enforcing algorithmic fairness~\citep{ChoiAAAI20,ChoiAAAI21}, making predictions under missing data~\citep{correia2020joints,khosravi2019tractable,LiUAI21}, data compression~\citep{LiuICLR22lossless}, and anomaly detection~\citep{dietrichstein2022anomaly}. 

Recent advancements in PC learning and regularization~\citep{shih2021hyperspns,LiuNeurIPS21}, and efficient implementations~\citep{peharz2020einsum,molina2019spflow,DangAAAI21} have been pushing the limits of PC's expressiveness and scalability such that they can even match the performance of less tractable deep generative models, including flow-based models and VAEs. 
However, the performance of PCs plateaus as model size increases. This suggests that to further boost the performance of PCs, simply scaling up the model size does not suffice and we need to better utilize the available capacity.

We discover that this might be caused by the fact that the capacity of large PCs is wasted. 
As shown in Figure~\ref{fig:params_hist}, most parameters in a PC with 2.18M parameters have close-to-zero values, which have little effect on the PC distribution.
Since existing PC structures usually have fully-connected
parameter layers~\citep{LiuNeurIPS21,rahman2014cutset}, this indicates that the parameter values are only sparsely used.

In this work, we propose to better exploit the sparsity of large PC models by two structure learning primitives --- \emph{pruning} and \emph{growing}. 
Specifically, the goal of the pruning operation is to identify and remove unimportant sub-networks of a PC. This is done by quantifying the importance of PC parameters \wrt a dataset using \emph{circuit flows}, a theoretically-grounded metric that upper bounds the drop of log-likelihood caused by pruning. Compared to L1 regularization, the proposed pruning operator is more informed by the PC semantics, and hence quantifies the global effects of pruning much more effectively.
Empirically, the proposed pruning method achieves a compression rate of 80-98\% with at most 1\% drop in likelihood on various PCs. 

The proposed growing operation increases the model size by copying its existing components and injecting noise. In particular, when applied to PCs compressed by the pruning operation, growing produces larger PCs that can be optimized to achieve better performance. Applying pruning and growing iteratively can greatly refine the structure and parameters of a PC. Empirically, the log-likelihoods metric can improve by 2\% to 10\% after a few iterations.
Compared to existing PC learners as well as less tractable deep generative models such as VAEs and flow-based models, our proposed method achieves state-of-the-art density estimation results on image datasets including
MNIST, EMNIST, FashionMNIST, and the Penn Tree Bank language modeling task.\footnote{Code and experiments are available at~ \url{https://github.com/UCLA-StarAI/SparsePC}.}

\section{Probabilistic Circuits}

\label{sec:pc}
\emph{Probabilistic circuits (PCs)}~\cite{PCTuto20,ProbCirc20} model probability distributions with a structured computation graph. They are an umbrella term for a large family of tractable probabilistic models including arithmetic circuits~\cite{darwiche02KR,darwicheJACM-POLY}, sum-product networks (SPNs)~\cite{poon2011sum}, cutset networks~\cite{rahman2014cutset}, and-or search spaces~\cite{marinescu2005and}, and probabilistic sentential decision diagrams~\cite{KisaVCD14}.
The syntax and semantics of PCs are defined as follows.

\begin{figure}[t]
\centering
\begin{subfigure}[b]{0.34\textwidth}
\centering
\scalebox{0.75}{\begin{tikzpicture}
    \varnode[line width=\midlinewidth]{z1}{$Z_1$}
    \varnode[line width=\midlinewidth,right=\midsmalldist of z1]{x1}{$X_1$}
    \varnode[line width=\midlinewidth,below=\midsmalldist of z1,xshift=-15pt]{z2}{$Z_2$}
    \varnode[line width=\midlinewidth,below=\midsmalldist of z1,xshift=15pt]{x3}{$X_3$}
    
    \upedge[line width=\midlinewidth] {z1} {z2}
    \upedge[line width=\midlinewidth] {z1} {x3}
    \upedge[line width=\midlinewidth] {z1} {x1}
    
    \varnode[line width=\midlinewidth,below=\midsmalldist of z2,xshift=15pt]{x4}{$X_4$}
    \varnode[line width=\midlinewidth,below=\midsmalldist of z2,xshift=-15pt]{x2}{$X_2$}
    
    \upedge[line width=\midlinewidth] {z2} {x4}
    \upedge[line width=\midlinewidth] {z2} {x2}
\end{tikzpicture}}
    \caption{\label{fig:pc1}}
\end{subfigure}
\hfill
\begin{subfigure}[b]{0.6\textwidth}
\centering
\scalebox{0.75}{ \begin{tikzpicture}
    \sumnode[line width=\midlinewidth]{s11}
    \annotatenode{s11}{$\color{red}{\tiny.12}$}{0pt}{13pt}
    
    \prodnode[line width=\midlinewidth,left=24pt of s11,yshift=18pt]{p11}
    \annotatenode{p11}{$\color{red}{\tiny.279}$}{0pt}{13pt}
    \prodnode[line width=\midlinewidth,left=24pt of s11,yshift=-18pt]{p12}
    \annotatenode{p12}{$\color{red}{\tiny.014}$}{0pt}{13pt}
    
    \colorweigedge[line width=\midlinewidth]{s11}{p11}{$0.4$}{black}{white}{black}
    \colorweigedge[line width=\midlinewidth]{s11}{p12}{$0.6$}{black}{white}{black}

    \sumnode[line width=\midlinewidth,left=28pt of p11]{s21}
    \annotatenode{s21}{$\color{red}{\tiny.388}$}{0pt}{13pt}
    \sumnode[line width=\midlinewidth,left=28pt of p12]{s22}
    \annotatenode{s22}{$\color{red}{\tiny.066}$}{0pt}{13pt}
    \edge[line width=\midlinewidth]{p11}{s21}å
    \edge[line width=\midlinewidth]{p12}{s22}
    
    \bernode[line width=\midlinewidth,left=-18pt of p11,yshift=29pt]{v11}{}%
    \bernode[line width=\midlinewidth,left=28pt of p11,yshift=29pt]{v31}{}%
    \edge[line width=\midlinewidth]{p11}{v11}
    \edge[line width=\midlinewidth]{p11}{v31}
    \annotatenode{v11}{$X_1\!\sim\!\mathcal{B}(.1)$}{0pt}{15pt}
    \annotatenode{v31}{$X_3\!\sim\!\mathcal{B}(.2)$}{0pt}{15pt}
    \annotatenode{v11}{$\color{red}{\tiny.9}$}{-14pt}{0pt}
    \annotatenode{v31}{$\color{red}{\tiny.8}$}{-14pt}{0pt}
    
    \bernode[line width=\midlinewidth,left=28pt of p12,yshift=-25pt]{v32}{}%
    \bernode[line width=\midlinewidth,left=-18pt of p12,yshift=-25pt]{v12}{}%
    \edge[line width=\midlinewidth]{p12}{v12}
    \edge[line width=\midlinewidth]{p12}{v32}
    \annotatenode{v12}{$X_1\!\sim\!\mathcal{B}(.7)$}{0pt}{-15pt}
    \annotatenode{v32}{$X_3\!\sim\! \mathcal{B}(.3)$}{0pt}{-15pt}
    \annotatenode{v12}{$\color{red}{\tiny.3}$}{-14pt}{0pt}
    \annotatenode{v32}{$\color{red}{\tiny.7}$}{-14pt}{0pt}
    
    \prodnode[line width=\midlinewidth,left=24pt of s21, yshift=25pt]{p21}
    \annotatenode{p21}{$\color{red}{\tiny.48}$}{0pt}{13pt}
    \prodnode[line width=\midlinewidth,left=24pt of s22, yshift=-25pt]{p22}
    \annotatenode{p22}{$\color{red}{\tiny.02}$}{0pt}{13pt}
    \colorweigedge[line width=\midlinewidth]{s21}{p21}{$0.8$}{black}{white}{black}
    \colorweigedge[line width=\midlinewidth]{s21}{p22}{$0.2$}{black}{white}{black}
    \colorweigedge[line width=\midlinewidth]{s22}{p21}{$0.1$}{black}{white}{black}
    \colorweigedge[line width=\midlinewidth]{s22}{p22}{$0.9$}{black}{white}{black}

    \bernode[line width=\midlinewidth,left=18pt of p21,yshift=0pt]{v21}{}%
    \bernode[line width=\midlinewidth,left=18pt of p21,yshift=-29pt]{v22}{}%
    \annotatenode{v21}{$X_2\!\sim\!\mathcal{B}(.6)$}{-40pt}{0pt}
    \annotatenode{v22}{$X_4\!\sim\! \mathcal{B}(.8)$}{-40pt}{0pt}
    \annotatenode{v21}{$\color{red}{\tiny.6}$}{0pt}{13pt}
    \annotatenode{v22}{$\color{red}{\tiny.8}$}{0pt}{13pt}
    \edge[line width=\midlinewidth]{p21}{v21}
    \edge[line width=\midlinewidth]{p21}{v22}

    \bernode[line width=\midlinewidth,left=18pt of p22,yshift=0pt]{v42}{}%
    \bernode[line width=\midlinewidth,left=18pt of p22,yshift=29pt]{v41}{}%
    \annotatenode{v41}{$X_2\!\sim\! \mathcal{B}(.1)$}{-40pt}{0pt}
    \annotatenode{v41}{$\color{red}{\tiny.1}$}{0pt}{13pt}
    \annotatenode{v42}{$X_4\!\sim\!\mathcal{B}(.2)$}{-40pt}{0pt}
     \annotatenode{v42}{$\color{red}{\tiny.2}$}{0pt}{13pt}
    \edge[line width=\midlinewidth]{p22}{v41}
    \edge[line width=\midlinewidth]{p22}{v42}
\end{tikzpicture}}
    \caption{\label{fig:pc2}}
\end{subfigure}

\caption{\label{fig:pc}A smooth and decomposable PC~(\subref{fig:pc2}) and an equivalent Bayesian network~(\subref{fig:pc1}). The Bayesian network is over 4 variables $\X=\{X_{1},X_{2},X_{3},X_{4}\}$ and 2 hidden variables $\Z=\{Z_{1},Z_{2}\}$ with $h=2$ hidden states. The feedforward computation order is from left to right; $\bigodot$ are input Bernoulli distributions, $\bigotimes$ are product units, and $\bigoplus$ are sum units; parameter values are annotated in the box. The probability of each unit given input assignment 
$\{X_1\!=\!0,X_2\!=\!1,X_3\!=\!0,X_4\!=\!1\}$ 
is labeled~red.}

\end{figure}
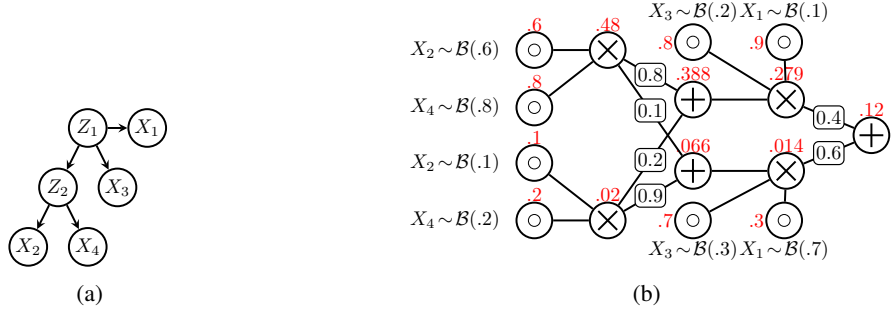

\begin{defn}[Probabilistic Circuit]
A PC $\PC\!:=\!(\graph,\params)$ represents a joint probability distribution $\p(\X)$ over random variables $\X$ through a directed acyclic (computation) graph (DAG) $\graph$ parameterized by $\params$. Similar to neural networks, each node in the DAG defines a computational unit. Specifically, the DAG $\graph$ consists of three types of units --- \emph{input}, \emph{sum}, and \emph{product}. Every leaf node in $\graph$ is an input unit; every inner unit $n$ (i.e., sum or product) receives \emph{inputs} from its children $\ch(n)$, and computes \emph{output}, which encodes a probability distribution $\p_n$ defined recursively as follows:
\begin{equation}
{\p}_n(\x):=
\begin{cases}
    f_n(\x) &\text{if $n$ is an input unit,} \\
    \prod_{c\in\ch(n)} \p_c(\x) &\text{if $n$ is a product unit,} \\
    \sum_{c\in\ch(n)} \theta_{c \given n}\cdot \p_c(\x) &\text{if $n$ is a sum unit,}
\end{cases}
\label{eq:EVI}
\end{equation}
\noindent where $f_n(\x)$ is a univariate input distribution (e.g, Gaussian, Categorical), and $\theta_{c \given n}$ denotes the parameter that corresponds to edge $(n,c)$ in the DAG. For every sum unit $n$, its input parameters sum up to one, \ie $\sum_{c \in \ch(n)} \theta_{c \given n} = 1$.
Intuitively, a product unit defines a factorized distribution over its inputs, and a sum unit represents a mixture over its input distributions with weights $\{\theta_{c \given n}\!:\!c \in \ch(n)\}$. Finally, the probability distribution of a PC (\ie $\p_{\PC}$) is defined as the distribution represented by its root unit $r$ (\ie $\p_r(\x)$), that is, its output neuron. The size of a PC, denoted $\abs{\PC}=\abs{\params}$, is the number of parameters in $\PC$. 
We assume w.l.o.g. that a PC alternates between layers of sum and product units before reaching its inputs. Figure~\ref{fig:pc} shows an example of a PC.
\end{defn}

Computing the (log)likelihood of a PC $\PC$ given a sample $\x$ is equivalent to evaluating its computation units in $\graph$ in a feedforward manner following Equation~\ref{eq:EVI}.
The key property that separates PCs from other deep probabilistic models such as flows~\citep{dinh2014nice} and VAEs~\citep{kingma2013auto} is their \emph{tractability}, which is the ability to exactly and efficiently answer various probabilistic queries. This paper focuses on PCs that support linear time (\wrt model size) marginal probability computation, as they are increasingly used in downstream applications such as data compression~\citep{LiuICLR22lossless} and making predictions under missing data~\citep{khosravi2019tractable}, and also achieve on-par expressiveness~\citep{LiuICLR22lossless,LiuNeurIPS21,LiangUAI17}. To support efficient marginal inference, PCs need to be \emph{smooth} and \emph{decomposable}.

\begin{defn}[Smoothness and Decomposability~\citep{darwiche2002knowledge}]
The \emph{scope} $\phi(n)$ of a PC unit $n$ is the set of input variables that it depends on; then, 
(1) a product unit is \emph{decomposable} if its children have disjoint scope;
(2) a sum unit is \emph{smooth} if its children have identical scope.
A PC is decomposable if all of its product units are decomposable; a PC is smooth if all of its sum units are smooth.
\end{defn}

Decomposability ensures that every product unit encodes a well-defined factorized distribution over disjoint sets of variables; smoothness ensures that the mixture components of every sum units are well-defined over the same set of variables. 
Both structural properties will be the key to guaranteeing the effectiveness of the structure learning algorithms proposed in the following sections.

\section{Probabilistic Circuit Model Compression via Pruning}
\label{sec:prune}

\begin{figure}[t]
\centering
\begin{subfigure}[b]{0.328\columnwidth}
\centering
\scalebox{0.6}{ \begin{tikzpicture}
\sumnode[line width=\midlinewidth]{s11}
    \sumnode[line width=\midlinewidth,right=15pt of s11]{s12}
    \sumnode[line width=\midlinewidth,right=15pt of s12]{s13}
    
    \node[circle, above=12pt of s11] (r11){};
    \node[circle, above=12pt of s12] (r12){};
    \node[circle, above=12pt of s13] (r13){};
    \cedge[line width=\midlinewidth]{r11}{s11}
    \cedge[line width=\midlinewidth]{r12}{s12}
    \cedge[line width=\midlinewidth]{r13}{s13}
    
    \prodnode[line width=\midlinewidth,below=23pt of s11]{p11}
    \prodnode[line width=\midlinewidth,below=23pt of s12]{p12}
    \prodnode[line width=\midlinewidth,below=23pt of s13]{p13}
    
    \cedge[line width=\midlinewidth]{s11}{p11}
    \cedge[line width=\midlinewidth]{s11}{p12}
    \cedge[line width=\midlinewidth,draw=red]{s11}{p13}
    \cedge[line width=\midlinewidth,draw=red]{s12}{p11}
    \cedge[line width=\midlinewidth]{s12}{p12}
    \cedge[line width=\midlinewidth, draw=red]{s12}{p13}
    \cedge[line width=\midlinewidth]{s13}{p11}
    \cedge[line width=\midlinewidth, draw=red]{s13}{p12}
    \cedge[line width=\midlinewidth, draw=red]{s13}{p13}

    \varnode[line width=\midlinewidth,below=18pt of p11,xshift=-45pt]{v11}{}
    \varnode[line width=\midlinewidth,below=18pt of p12,xshift=-45pt]{v12}{}
    \varnode[line width=\midlinewidth,below=18pt of p13,xshift=-45pt]{v13}{}
    
    \varnode[line width=\midlinewidth,below=18pt of p11,xshift=45pt]{v11_2}{}
    \varnode[line width=\midlinewidth,below=18pt of p12,xshift=45pt]{v12_2}{}
    \varnode[line width=\midlinewidth,below=18pt of p13,xshift=45pt]{v13_2}{}
    
    \cedge[line width=\midlinewidth]{p11}{v11,v11_2}
    \cedge[line width=\midlinewidth]{p12}{v12,v12_2}
    \cedge[line width=\midlinewidth]{p13}{v13,v13_2}
\end{tikzpicture}}
    \caption{\label{fig:op1}PC with fully connected layers}
\end{subfigure}
\begin{subfigure}[b]{0.328\columnwidth}
\centering
\scalebox{0.6}{ \begin{tikzpicture}

    \sumnode[line width=\midlinewidth, right=180pt of s11]{s21}
    \sumnode[line width=\midlinewidth,right=15pt of s21]{s22}
    \sumnode[line width=\midlinewidth,right=15pt of s22]{s23}
    
    \node[circle, above=12pt of s21] (r21){};
    \node[circle, above=12pt of s22] (r22){};
    \node[circle, above=12pt of s23] (r23){};
    \cedge[line width=\midlinewidth]{r21}{s21}
    \cedge[line width=\midlinewidth]{r22}{s22}
    \cedge[line width=\midlinewidth]{r23}{s23}
    
    \prodnode[line width=\midlinewidth,below=23pt of s21, xshift=20pt]{p21}
    \prodnode[line width=\midlinewidth,below=23pt of s22, xshift=20pt]{p22}

    \cedge[line width=\midlinewidth]{s21}{p21}
    \cedge[line width=\midlinewidth]{s21}{p22}
    \cedge[line width=\midlinewidth]{s22}{p22}
    \cedge[line width=\midlinewidth]{s23}{p21}

    \varnode[line width=\midlinewidth,below=18pt of p21,xshift=-30pt]{v21}{}
    \varnode[line width=\midlinewidth,below=18pt of p22,xshift=-30pt]{v22}{}
    
    \varnode[line width=\midlinewidth,below=18pt of p21,xshift=30pt]{v211}{}
    \varnode[line width=\midlinewidth,below=18pt of p22,xshift=30pt]{v221}{}
    
    \cedge[line width=\midlinewidth]{p21}{v21,v211}
    \cedge[line width=\midlinewidth]{p22}{v22,v221}

\end{tikzpicture}}
    \caption{\label{fig:op2}PC after pruning operation}
\end{subfigure}
\begin{subfigure}[b]{0.328\columnwidth}
\centering
\scalebox{0.6}{ \begin{tikzpicture}
    \sumnode[line width=\midlinewidth, right=150pt of s21]{s31}
    \sumnode[line width=\midlinewidth,right=15pt of s31]{s32}
    \sumnode[line width=\midlinewidth,right=15pt of s32]{s33}
    
    \sumnode[line width=\midlinewidth, right=15pt of s33, draw=gold2]{s31n}
    \sumnode[line width=\midlinewidth, right=15pt of s31n, draw=gold2]{s32n}
    \sumnode[line width=\midlinewidth, right=15pt of s32n, draw=gold2]{s33n}
    
    \node[circle, above=12pt of s31] (r31){};
    \node[circle, above=12pt of s32] (r32){};
    \node[circle, above=12pt of s33] (r33){};
    
    \node[circle, above=12pt of s31n] (r31n){};
    \node[circle, above=12pt of s32n] (r32n){};
    \node[circle, above=12pt of s33n] (r33n){};
    \cedge[line width=\midlinewidth]{r31}{s31}
    \cedge[line width=\midlinewidth]{r32}{s32}
    \cedge[line width=\midlinewidth]{r33}{s33}
    
    \cedge[line width=\midlinewidth, draw=gold2]{r31n}{s31n}
    \cedge[line width=\midlinewidth, draw=gold2]{r32n}{s32n}
    \cedge[line width=\midlinewidth, draw=gold2]{r33n}{s33n}
    
    \prodnode[line width=\midlinewidth,below=23pt of s31, xshift=20pt]{p31}
    \prodnode[line width=\midlinewidth,below=23pt of s32, xshift=20pt]{p32}
    \prodnode[line width=\midlinewidth,below=23pt of s31n, xshift=20pt,draw=gold2]{p31n}
    \prodnode[line width=\midlinewidth,below=23pt of s32n, xshift=20pt,draw=gold2]{p32n}

    \cedge[line width=\midlinewidth]{s31}{p31}
    \cedge[line width=\midlinewidth]{s31}{p32}
    \cedge[line width=\midlinewidth]{s32}{p32}
    \cedge[line width=\midlinewidth]{s33}{p31}
    
    \cedge[line width=\midlinewidth,draw=gold2]{s31n}{p31n}
    \cedge[line width=\midlinewidth,draw=gold2]{s31n}{p32n}
    \cedge[line width=\midlinewidth,draw=gold2]{s32n}{p32n}
    \cedge[line width=\midlinewidth,draw=gold2]{s33n}{p31n}
    
    \cedge[line width=\midlinewidth,draw=petroil2]{s31}{p31n}
    \cedge[line width=\midlinewidth,draw=petroil2]{s31}{p32n}
    \cedge[line width=\midlinewidth,draw=petroil2]{s32}{p32n}
    \cedge[line width=\midlinewidth,draw=petroil2]{s33}{p31n}
    
    \cedge[line width=\midlinewidth,draw=violet]{s31n}{p31}
    \cedge[line width=\midlinewidth,draw=violet]{s31n}{p32}
    \cedge[line width=\midlinewidth,draw=violet]{s32n}{p32}
    \cedge[line width=\midlinewidth,draw=violet]{s33n}{p31}

    \varnode[line width=\midlinewidth,below=18pt of p31,xshift=-23pt]{v31}{}
    \varnode[line width=\midlinewidth,below=18pt of p32,xshift=-30pt]{v32}{}
    
    \varnode[line width=\midlinewidth,below=18pt of p31,xshift=30pt]{v311}{}
    \varnode[line width=\midlinewidth,below=18pt of p32,xshift=23pt]{v321}{}

    \cedge[line width=\midlinewidth]{p31}{v31,v311}
    \cedge[line width=\midlinewidth]{p32}{v32,v321}

    \varnode[line width=\midlinewidth,below=18pt of p31n,xshift=-23pt,draw=gold2]{v31n}{}
    \varnode[line width=\midlinewidth,below=18pt of p32n,xshift=-30pt,draw=gold2]{v32n}{}
    
    \varnode[line width=\midlinewidth,below=18pt of p31n,xshift=30pt,draw=gold2]{v311n}{}
    \varnode[line width=\midlinewidth,below=18pt of p32n,xshift=23pt,draw=gold2]{v321n}{}
    
    \cedge[line width=\midlinewidth, draw=gold2]{p31n}{v31n,v311n}
    \cedge[line width=\midlinewidth, draw=gold2]{p32n}{v32n,v321n}
\end{tikzpicture}}
    \caption{\label{fig:op3-first}PC after growing operation}
\end{subfigure}
\caption{\label{fig:op}A demonstration of the pruning and growing operation. From~\ref{fig:op1} to~\ref{fig:op2}, the red edges are pruned. From~\ref{fig:op2} to~\ref{fig:op3-first}, the nodes are doubled, and each parameter is copied 3 times.
}
\end{figure}
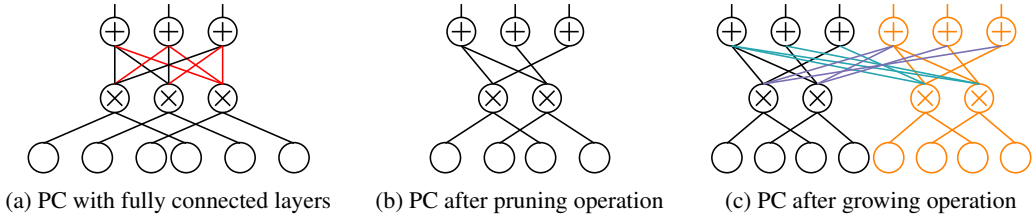

Figure~\ref{fig:params_hist} shows that most parameters in a large PC are very close to zero. Given that these parameters are weights associated with mixture (sum unit) components, the corresponding edges and sub-circuits have little impact on the sum unit output. Hence, by pruning away these unimportant components, it is possible to significantly reduce model size while retaining model expressiveness.
Figure~\ref{fig:op2} illustrates the result of pruning five (red) edges from the PC in Figure~\ref{fig:op1}. Given a PC and a dataset, our goal is to efficiently identify a set of edges to prune, such that the log-likelihood gap between the pruned PC 
and the original PC on the given dataset is minimized.

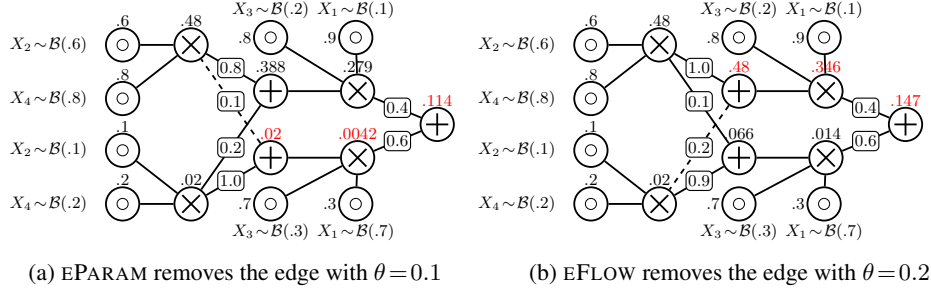
\begin{figure}
\footnotesize 
    \centering
    \begin{subfigure}[b]{0.44\textwidth}
\scalebox{0.70}{\begin{tikzpicture}
    \sumnode[line width=\midlinewidth]{s11}
    \annotatenode{s11}{$\color{red}{\tiny.114}$}{0pt}{13pt}
    
    \prodnode[line width=\midlinewidth,left=24pt of s11,yshift=18pt]{p11}
    \annotatenode{p11}{$\color{black}{\tiny.279}$}{0pt}{13pt}
    \prodnode[line width=\midlinewidth,left=24pt of s11,yshift=-18pt]{p12}
    \annotatenode{p12}{$\color{red}{\tiny.0042}$}{0pt}{13pt}
    
    \colorweigedge[line width=\midlinewidth]{s11}{p11}{$0.4$}{black}{white}{black}
    \colorweigedge[line width=\midlinewidth]{s11}{p12}{$0.6$}{black}{white}{black}

    \sumnode[line width=\midlinewidth,left=28pt of p11]{s21}
    \annotatenode{s21}{$\color{black}{\tiny.388}$}{0pt}{13pt}
    \sumnode[line width=\midlinewidth,left=28pt of p12]{s22}
    \annotatenode{s22}{$\color{red}{\tiny.02}$}{0pt}{13pt}
    \edge[line width=\midlinewidth]{p11}{s21}
    \edge[line width=\midlinewidth]{p12}{s22}
    
    \bernode[line width=\midlinewidth,left=-18pt of p11,yshift=29pt]{v11}{}%
    \bernode[line width=\midlinewidth,left=28pt of p11,yshift=29pt]{v31}{}%
    \edge[line width=\midlinewidth]{p11}{v11}
    \edge[line width=\midlinewidth]{p11}{v31}
    \annotatenode{v11}{$X_1\!\sim\!\mathcal{B}(.1)$}{0pt}{15pt}
    \annotatenode{v31}{$X_3\!\sim\!\mathcal{B}(.2)$}{0pt}{15pt}
    \annotatenode{v11}{$\color{black}{\tiny.9}$}{-14pt}{0pt}
    \annotatenode{v31}{$\color{black}{\tiny.8}$}{-14pt}{0pt}
    
    \bernode[line width=\midlinewidth,left=28pt of p12,yshift=-25pt]{v32}{}%
    \bernode[line width=\midlinewidth,left=-18pt of p12,yshift=-25pt]{v12}{}%
    \edge[line width=\midlinewidth]{p12}{v12}
    \edge[line width=\midlinewidth]{p12}{v32}
    \annotatenode{v12}{$X_1\!\sim\!\mathcal{B}(.7)$}{0pt}{-15pt}
    \annotatenode{v32}{$X_3\!\sim\! \mathcal{B}(.3)$}{0pt}{-15pt}
    \annotatenode{v12}{$\color{black}{\tiny.3}$}{-14pt}{0pt}
    \annotatenode{v32}{$\color{black}{\tiny.7}$}{-14pt}{0pt}
    
    \prodnode[line width=\midlinewidth,left=24pt of s21, yshift=25pt]{p21}
    \annotatenode{p21}{$\color{black}{\tiny.48}$}{0pt}{13pt}
    \prodnode[line width=\midlinewidth,left=24pt of s22, yshift=-25pt]{p22}
    \annotatenode{p22}{$\color{black}{\tiny.02}$}{0pt}{13pt}
    \colorweigedge[line width=\midlinewidth]{s21}{p21}{$0.8$}{black}{white}{black}
    \colorweigedge[line width=\midlinewidth]{s21}{p22}{$0.2$}{black}{white}{black}
    \colorweigedge[line width=\midlinewidth,draw=black,dashed]{s22}{p21}{$0.1$}{black}{white}{black}
    \colorweigedge[line width=\midlinewidth,draw=black]{s22}{p22}{$1.0$}{black}{white}{black}

    \bernode[line width=\midlinewidth,left=18pt of p21,yshift=0pt]{v21}{}%
    \bernode[line width=\midlinewidth,left=18pt of p21,yshift=-29pt]{v22}{}%
    \annotatenode{v21}{$X_2\!\sim\!\mathcal{B}(.6)$}{-40pt}{0pt}
    \annotatenode{v22}{$X_4\!\sim\! \mathcal{B}(.8)$}{-40pt}{0pt}
    \annotatenode{v21}{$\color{black}{\tiny.6}$}{0pt}{13pt}
    \annotatenode{v22}{$\color{black}{\tiny.8}$}{0pt}{13pt}
    \edge[line width=\midlinewidth]{p21}{v21}
    \edge[line width=\midlinewidth]{p21}{v22}

    \bernode[line width=\midlinewidth,left=18pt of p22,yshift=0pt]{v42}{}%
    \bernode[line width=\midlinewidth,left=18pt of p22,yshift=29pt]{v41}{}%
    \annotatenode{v41}{$X_2\!\sim\! \mathcal{B}(.1)$}{-40pt}{0pt}
    \annotatenode{v41}{$\color{black}{\tiny.1}$}{0pt}{13pt}
    \annotatenode{v42}{$X_4\!\sim\!\mathcal{B}(.2)$}{-40pt}{0pt}
     \annotatenode{v42}{$\color{black}{\tiny.2}$}{0pt}{13pt}
    \edge[line width=\midlinewidth]{p22}{v41}
    \edge[line width=\midlinewidth]{p22}{v42}
\end{tikzpicture}}
    \caption{\label{fig:by_param}\eparam\ removes the edge with $\theta\!=\!0.1$}
\end{subfigure}
\begin{subfigure}[b]{0.49\textwidth}
\scalebox{0.70}{\begin{tikzpicture}
    \sumnode[line width=\midlinewidth]{s11}
    \annotatenode{s11}{$\color{red}{\tiny.147}$}{0pt}{13pt}
    
    \prodnode[line width=\midlinewidth,left=24pt of s11,yshift=18pt]{p11}
    \annotatenode{p11}{$\color{red}{\tiny.346}$}{0pt}{13pt}
    \prodnode[line width=\midlinewidth,left=24pt of s11,yshift=-18pt]{p12}
    \annotatenode{p12}{$\color{black}{\tiny.014}$}{0pt}{13pt}
    
    \colorweigedge[line width=\midlinewidth]{s11}{p11}{$0.4$}{black}{white}{black}
    \colorweigedge[line width=\midlinewidth]{s11}{p12}{$0.6$}{black}{white}{black}

    \sumnode[line width=\midlinewidth,left=28pt of p11]{s21}
    \annotatenode{s21}{$\color{red}{\tiny.48}$}{0pt}{13pt}
    \sumnode[line width=\midlinewidth,left=28pt of p12]{s22}
    \annotatenode{s22}{$\color{black}{\tiny.066}$}{0pt}{13pt}
    \edge[line width=\midlinewidth]{p11}{s21}
    \edge[line width=\midlinewidth]{p12}{s22}
    
    \bernode[line width=\midlinewidth,left=-18pt of p11,yshift=29pt]{v11}{}%
    \bernode[line width=\midlinewidth,left=28pt of p11,yshift=29pt]{v31}{}%
    \edge[line width=\midlinewidth]{p11}{v11}
    \edge[line width=\midlinewidth]{p11}{v31}
    \annotatenode{v11}{$X_1\!\sim\!\mathcal{B}(.1)$}{0pt}{15pt}
    \annotatenode{v31}{$X_3\!\sim\!\mathcal{B}(.2)$}{0pt}{15pt}
    \annotatenode{v11}{$\color{black}{\tiny.9}$}{-14pt}{0pt}
    \annotatenode{v31}{$\color{black}{\tiny.8}$}{-14pt}{0pt}
    
    \bernode[line width=\midlinewidth,left=28pt of p12,yshift=-25pt]{v32}{}%
    \bernode[line width=\midlinewidth,left=-18pt of p12,yshift=-25pt]{v12}{}%
    \edge[line width=\midlinewidth]{p12}{v12}
    \edge[line width=\midlinewidth]{p12}{v32}
    \annotatenode{v12}{$X_1\!\sim\!\mathcal{B}(.7)$}{0pt}{-15pt}
    \annotatenode{v32}{$X_3\!\sim\! \mathcal{B}(.3)$}{0pt}{-15pt}
    \annotatenode{v12}{$\color{black}{\tiny.3}$}{-14pt}{0pt}
    \annotatenode{v32}{$\color{black}{\tiny.7}$}{-14pt}{0pt}
    
    \prodnode[line width=\midlinewidth,left=24pt of s21, yshift=25pt]{p21}
    \annotatenode{p21}{$\color{black}{\tiny.48}$}{0pt}{13pt}
    \prodnode[line width=\midlinewidth,left=24pt of s22, yshift=-25pt]{p22}
    \annotatenode{p22}{$\color{black}{\tiny.02}$}{0pt}{13pt}
    \colorweigedge[line width=\midlinewidth]{s21}{p21}{$1.0$}{black}{white}{black}
    \colorweigedge[line width=\midlinewidth,dashed,draw=black]{s21}{p22}{$0.2$}{black}{white}{black}
    \colorweigedge[line width=\midlinewidth]{s22}{p21}{$0.1$}{black}{white}{black}
    \colorweigedge[line width=\midlinewidth]{s22}{p22}{$0.9$}{black}{white}{black}

    \bernode[line width=\midlinewidth,left=18pt of p21,yshift=0pt]{v21}{}%
    \bernode[line width=\midlinewidth,left=18pt of p21,yshift=-29pt]{v22}{}%
    \annotatenode{v21}{$X_2\!\sim\!\mathcal{B}(.6)$}{-40pt}{0pt}
    \annotatenode{v22}{$X_4\!\sim\! \mathcal{B}(.8)$}{-40pt}{0pt}
    \annotatenode{v21}{$\color{black}{\tiny.6}$}{0pt}{13pt}
    \annotatenode{v22}{$\color{black}{\tiny.8}$}{0pt}{13pt}
    \edge[line width=\midlinewidth]{p21}{v21}
    \edge[line width=\midlinewidth]{p21}{v22}

    \bernode[line width=\midlinewidth,left=18pt of p22,yshift=0pt]{v42}{}%
    \bernode[line width=\midlinewidth,left=18pt of p22,yshift=29pt]{v41}{}%
    \annotatenode{v41}{$X_2\!\sim\! \mathcal{B}(.1)$}{-40pt}{0pt}
    \annotatenode{v41}{$\color{black}{\tiny.1}$}{0pt}{13pt}
    \annotatenode{v42}{$X_4\!\sim\!\mathcal{B}(.2)$}{-40pt}{0pt}
     \annotatenode{v42}{$\color{black}{\tiny.2}$}{0pt}{13pt}
    \edge[line width=\midlinewidth]{p22}{v41}
    \edge[line width=\midlinewidth]{p22}{v42}
\end{tikzpicture}}
    \caption{\label{fig:by_flow}\eflow\ removes the edge with $\theta\!=\!0.2$}
\end{subfigure}
\caption{\label{fig:compare_heuristic_example}A case study comparing pruning heuristics (\eparam\ and \eflow) on the PC in Figure~\ref{fig:pc} given sample $\{X_1\!=\!0,X_2\!=\!1,X_3\!=\!0,X_4\!=\!1\}$. The pruned edges are dashed and parameters are re-normalized.
Compared to the likelihood computed in Figure~\ref{fig:pc}, the changed likelihoods are in red, showing that pruning by flows results in less likelihood decrease. 
}
\vspace{-1em}
\end{figure}

\paragraph{Pruning by parameters.} The parameter value statistics in Figure~\ref{fig:params_hist} suggest that a natural criterion is to prune edges by the magnitude of their corresponding parameter. This leads to the \eparam\ (edge parameters) heuristic, which selects the set of edges with the smallest parameters. However, edge parameters themselves are insufficient to quantify the importance of inputs to a sum unit in the entire PC's distribution. The parameters of a sum unit are normalized to be 1 so they only contain local information about the mixture components. 
Specifically, $\theta_{c \given n}$ merely defines the relative importance of edge $(n, c)$ in the conditional distribution represented by its corresponding sum unit $n$, not the joint distribution of the entire PC.
Figure~\ref{fig:by_param} illustrates what happens when the edge with the smallest parameter is pruned from the PC in Figure~\ref{fig:pc}.

However, as shown in Figure~\ref{fig:by_flow}, pruning another edge delivers better likelihoods as it accounts more for the ``global influence'' of edges on the PC's output. This global influence is highly related to the probabilistic ``circuit flow'' semantics of PCs. We will introduce circuit flows later in this section, along with their corresponding heuristics \eflow. Before that, we first introduce an intermediate concept based on the notion of generative significance of PCs.

\begin{figure}[t]
    \input{inputs/alg-sample}
\end{figure}
\paragraph{Pruning by generative significance.} A more informed pruning strategy needs to consider the global impact of edges on the distribution represented by the output of the PC. To achieve this, 
instead of viewing the distribution $\p_{\PC}$ in a feedforward manner following Equation~\ref{eq:EVI}, 
we quantify the significance of a unit or edge by the probability that it will be ``activated'' when drawing samples from the PC. 
Indeed, if the presence of an edge is hardly ever relevant to the generative sampling process, removing it will not significantly affect the PC's distribution.

Algorithm~\ref{alg:sample} shows how to draw samples from a PC distribution through a recursive implementation:  (1) for an input unit $n$ defined on variable $X$ (line~\ref{code-line:input}), the algorithm randomly samples value $x$ according to its input univariate distribution; (2) for a product unit (line~\ref{code-line:product}), by decomposability its children have disjoint scope, thus we draw samples from all input units and then concatenate the samples together; (3) for a sum unit $n$ (line~\ref{code-line:sum}), by smoothness its children have identical scope, thus we first randomly sample one of its input units according to the categorical distribution defined by sum parameters $\{\theta_{c \given n}:c \in \ch(n)\}$, and then sample from this input unit recursively. Besides actually drawing samples from the PC, we can also compute the probability that $n$ will be visited during the sampling process. This provides a good measure of the importance of unit $n$ to the PC distribution as a whole, which we define as the \emph{top-down probability}.

\begin{defn}[Top-down Probability]
\label{def:td-prob}
The top-down probability of each unit $n$ in a PC with parameters $\params$ is defined recursively as follows, assuming alternating sum and product layers: 
\begin{equation*}
\q(n;\params):=
\begin{cases}
    1 &\text{if $n$ is the root unit,} \\
    \sum_{m\in\pa(n)}  \q(m;\params) &\text{if $n$ is a sum unit,} \\
    \sum_{m\in\pa(n)}  \theta_{n \given m} \cdot \q(m;\params) &\text{if $n$ is a product unit,}
\end{cases}
\end{equation*}
\noindent where $\pa(n)$ are the units that take $n$ as input in the feedforward computation. 
Moreover, the top-down probability of a sum edge $(n,c)$ is defined as $\q(n,c;\params) = \theta_{c \given n} \cdot \q(n;\params)$.
\end{defn}

The top-down probability of the root is always 1; a product unit passes its top-down probability to all its inputs, and a sum unit distributes its top-down probability to its inputs proportional to the corresponding edge weights. Therefore, the top-down probability of a non-root unit is summing over all probabilities it receives from its outputs. 

The top-down probability of all PC units and sum edges can be computed in a single backward pass over the PC's computation graph. Following the intuition that the top-down probability defines the probability that units will be visited during the sampling process, pruning edges with the smallest top-down probability constitutes a reasonable pruning strategy.

\paragraph{Pruning by circuit flows.} The top-down probability $\q(n;\params)$ represents the probability of reaching unit $n$ in an unconditional random sampling process. Despite its ability to capture global information of PC parameters, the top-down probability is not tailored to a specific dataset.
Therefore, to further utilize the dataset information, we can measure the probability of reaching certain units/edges in the sampling process \emph{conditioning on some instance $\x$ being sampled}. 
To bridge this gap, we define circuit flow as a sample-dependent version of the top-down probability.

\begin{defn}[Circuit Flow\footnote{Earlier work defined ``circuit flow'' or ``expected circuit flow'' in the context of parameter learning~\citep{ChoiAAAI21,LiuNeurIPS21,DangPGM20}, without observing the connection to sampling. We contribute its more intuitive sampling semantics here.}]
\label{def:flow}
For a given PC with parameters $\params$ and example $\x$, the circuit flow of unit $n$ on example $\x$ is the probability that $n$ will be visited during the sampling procedure conditioned on $\x$ being sampled. This can be computed recursively as follows, assuming alternating sum and product layers:
\begin{equation*}
\flow_{n} (\x) =
\begin{cases}
    1 &\text{if $n$ is the root unit,} \\
    \sum_{m\in\pa(n)}  \flow_m(\x) &\text{if $n$ is a sum unit,} \\
    \sum_{m\in\pa(n)}  \frac{\theta_{n \given m} \cdot \p_{n} (\x)}{\p_{m} (\x)} \cdot \flow_m(\x) &\text{if $n$ is a product unit.}
\end{cases}
\end{equation*}
Similarly, the edge flow $\flow_{n,c} (\x)$ on sample $\x$ is defined by $\flow_{n,c} (\x) =\theta_{c \given n} \cdot \p_{c} (\x)/\p_{n} (\x) \cdot \flow_n(\x)$. We further define $\flow_{n,c}(\data) = \sum_{\x \in \data} \flow_{n,c}(\x)$ as the \emph{aggregate edge flow} %
over dataset $\data$.
\end{defn}
Effectively, we can think of $\theta_{n\given m}^{\x}:=\theta_{n \given m} \cdot \p_{n} (\x)/\p_{m} (\x)$ as the posterior probability of component~$n$ in the mixture of sum unit $m$ \emph{conditioned on observing sample $\x$}. Then, circuit flow is the top-down probability under this $\params^{\x}$ reparameterization of the circuit: 
$\flow_{n} (\x) = \q(n;\params^{\x})$ and $\flow_{n,c} (\x)=\q(n,c;\params^{\x})$.

Circuit flow $\flow_{n}(\x)$ defines the probability of reaching unit $n$ in the top-down sampling procedure of Algorithm~\ref{alg:sample}, given that the sampled instance is $\x$. Therefore, edge flow $\flow_{n,c}(\x)$ is a natural metric of the importance of edge $(n,c)$ given $\x$. 
Intuitively, the aggregate circuit flow measures how many expected samples ``flow'' through certain edges.
We write $\eflow$ to refer to the heuristic that prunes edges with the smallest aggregate circuit flow. 
\begin{figure}[t]
    \centering
    \begin{subfigure}[b]{0.51\columnwidth}
    \centering
    \includegraphics[width=0.65\linewidth]{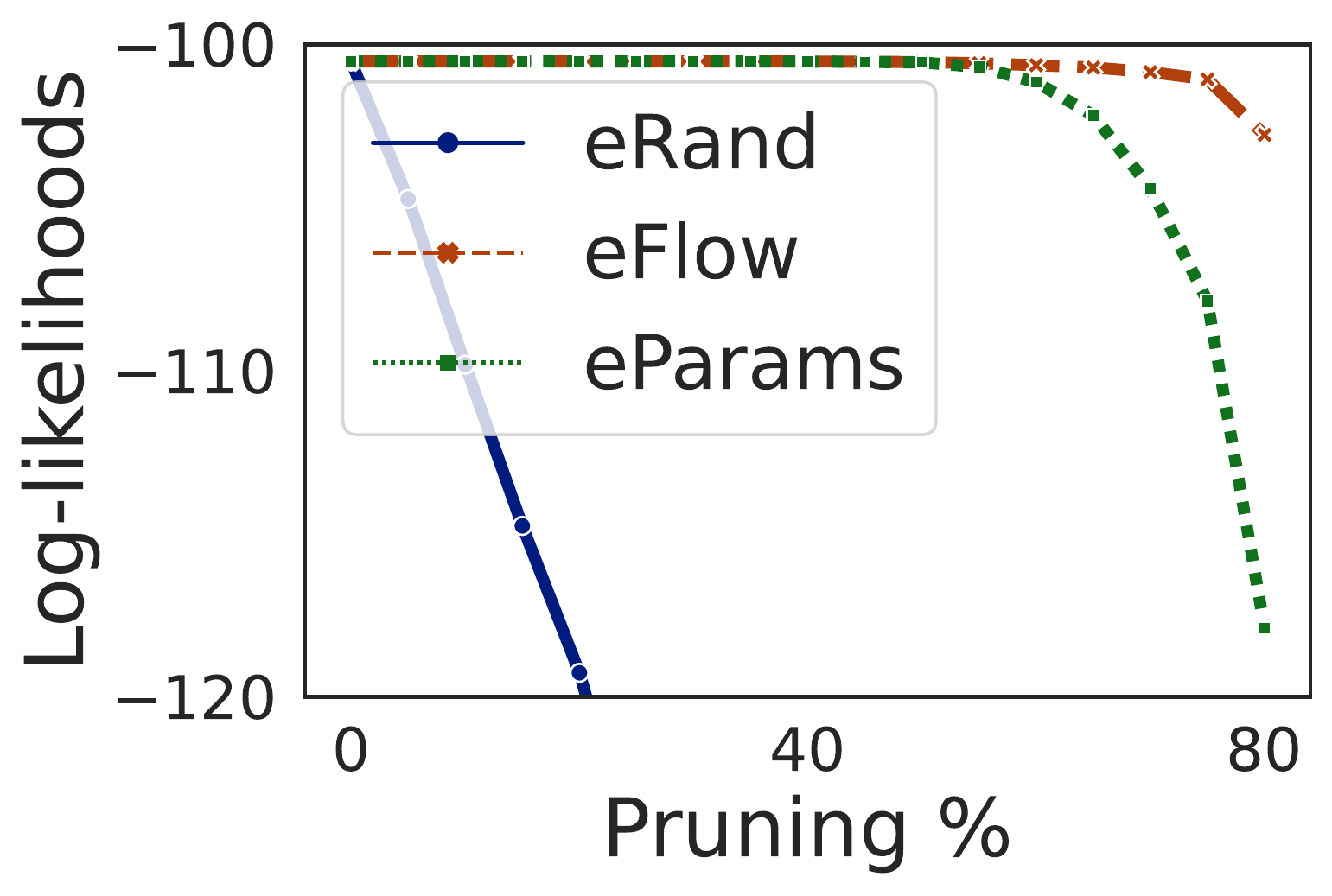}
    \caption{\label{fig:prune_heuristic} Comparison of heuristics $\erand$, $\eparam$, and $\eflow$. Heuristic $\eflow$ can prune up to 80\% of the parameters without much loglikelihoods decrease.}
    \end{subfigure}\hfill
    \begin{subfigure}[b]{0.46\columnwidth}
    \centering
    \includegraphics[width=0.72\linewidth]{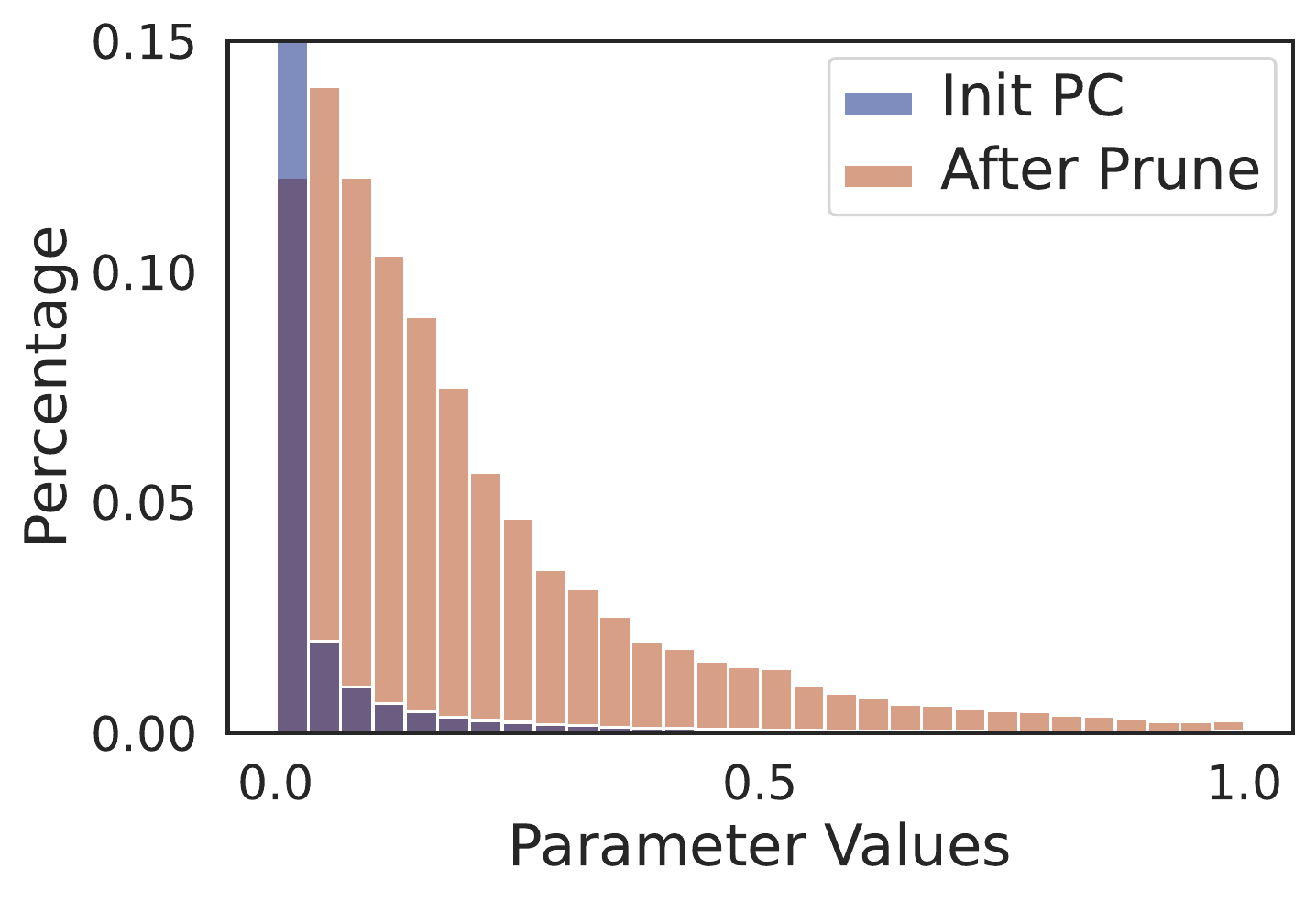}
    \caption{\label{fig:params_hist_after_prune} Histogram of parameters before (the same as in Figure~\ref{fig:params_hist}) and after pruning. The parameter values take higher significance after pruning.}
    \end{subfigure}
    \caption{\label{fig:prune-eval}Empirical evaluation of the pruning operation.}
\end{figure}
\paragraph{Empirical Analysis.}
Figure~\ref{fig:prune_heuristic} compares the effect of pruning heuristics \eparam, \eflow, as well as an uninformed strategy, prune randomly, which we denote as \erand. It shows that both \eparam\ and \eflow\ are reasonable pruning strategy, however, as we increase the percentage of pruned parameters, \eflow\ has less log-likelihoods drop compared with \eparam. Using \eflow\ heuristics we can pruning up to 80\% of the parameters without much log-likelihoods drop. As shown in Figure~\ref{fig:params_hist_after_prune}, the parameter distribution is more balanced after pruning compared to Figure~\ref{fig:params_hist}, indicating a higher significance of each edge. Section~\ref{sec:exp} will provide more empirical results. Before that, we first theoretically verify the effectiveness of the \eflow\ heuristic in the next section.

\section{Bounding and Approximating the Loss of Likelihood}

In this section, we theoretically quantify the impact of edge pruning on model performance. In particular, we establish an upper bound on the log-likelihood drop $\Delta\LL$ on a given dataset $\data$ by comparing (i) the original PC $\PC$ and (ii) the pruned PC $\PC_{\backslash \edges}$ caused by pruning away edges $\edges$:
\begin{equation}
    \Delta\LL(\data,\PC,\edges)\!=\! \LL(\data,\PC) - \LL(\data,\PC_{\backslash \edges}).
    \label{eq:prune-ll-opt}
\end{equation}
We start from the case of pruning one edge (\ie $\abs{\edges}\!=\!1$ in Equation~\ref{eq:prune-ll-opt}). In this case, the loss of likelihood can be quantified exactly using flows and edge parameters:

\begin{thm}[Log-likelihood drop of pruning one edge]
\label{thm:prune_one_ll}
For a PC $\PC$ and a dataset $\data$, the loss of log-likelihood by pruning away edge $(n,c)$ is
    \begin{align*}
        \Delta\LL(\data,\PC,\{(n,c)\})
        \!= \!\frac{1}{|\data|}\!\!\sum_{\x\in\data}\! \log\!\left(\frac{1-\theta_{c\given n}}{1 \!-\! \theta_{c\given n} \!+\! \theta_{c\given n} \flow_n(\x) \!-\! \flow_{n,c}(\x)}\!\right) 
        \!\leq\! \frac{-1}{|\data|}\!\sum_{\x\in\data}\log(1 \!-\! \flow_{n,c}(\x)).
    \end{align*}
\end{thm}
See proof in Appendix~\ref{app-sec:proof-single}.
By computing the second term in Theorem~\ref{thm:prune_one_ll}, we can pick the edge with the smallest log-likelihood drop. 
Additionally, the third term characterizes the log-likelihood drop without re-normalizing parameters of $\theta_{\cdot\given n}$. It suggests pruning the edge with the smallest edge flow.
A key insight from Theorem~\ref{thm:prune_one_ll} is that the log-likelihood drop 
depends explicitly on the edge flow $\flow_{n,c}(\x)$ and unit flow $\flow_{n}(\x)$.
This matches the intuition from Section~\ref{sec:prune} and suggests that the circuit flow heuristic proposed in the previous section is a good approximation of the derived upper~bound.

Next, we bound the log-likelihood drop of pruning multiple edges.

\begin{thm}[Log-likelihood drop of pruning multiple edges]
\label{thm:prune-multi}
Let $\PC$ be a PC and $\data$ be a dataset. 
For any set of edges $\edges$ in $\PC$,
if $\forall \x \!\in\! \data, \sum_{(n, c) \in \calE} \flow_{n,c}(\x) < 1$, the log-likelihood drop by pruning away $\edges$ is bounded and approximated by
\begin{align}
\begin{split}
    \Delta\LL(\data,\PC,\edges)
    \leq - \frac{1}{\abs{\data}}\sum_{\x}\log(1-\sum_{(n, c) \in \calE} \flow_{n,c}(\x)) 
    \approx \frac{1}{\abs{\data}}\sum_{(n, c) \in \calE} \flow_{n,c}(\data).
    \label{eq:prune_mul_d_ll}
\end{split}
\end{align}
\end{thm}

\begin{wrapfigure}{r}{0.41\textwidth}
    \centering
    \includegraphics[width=0.35\columnwidth]{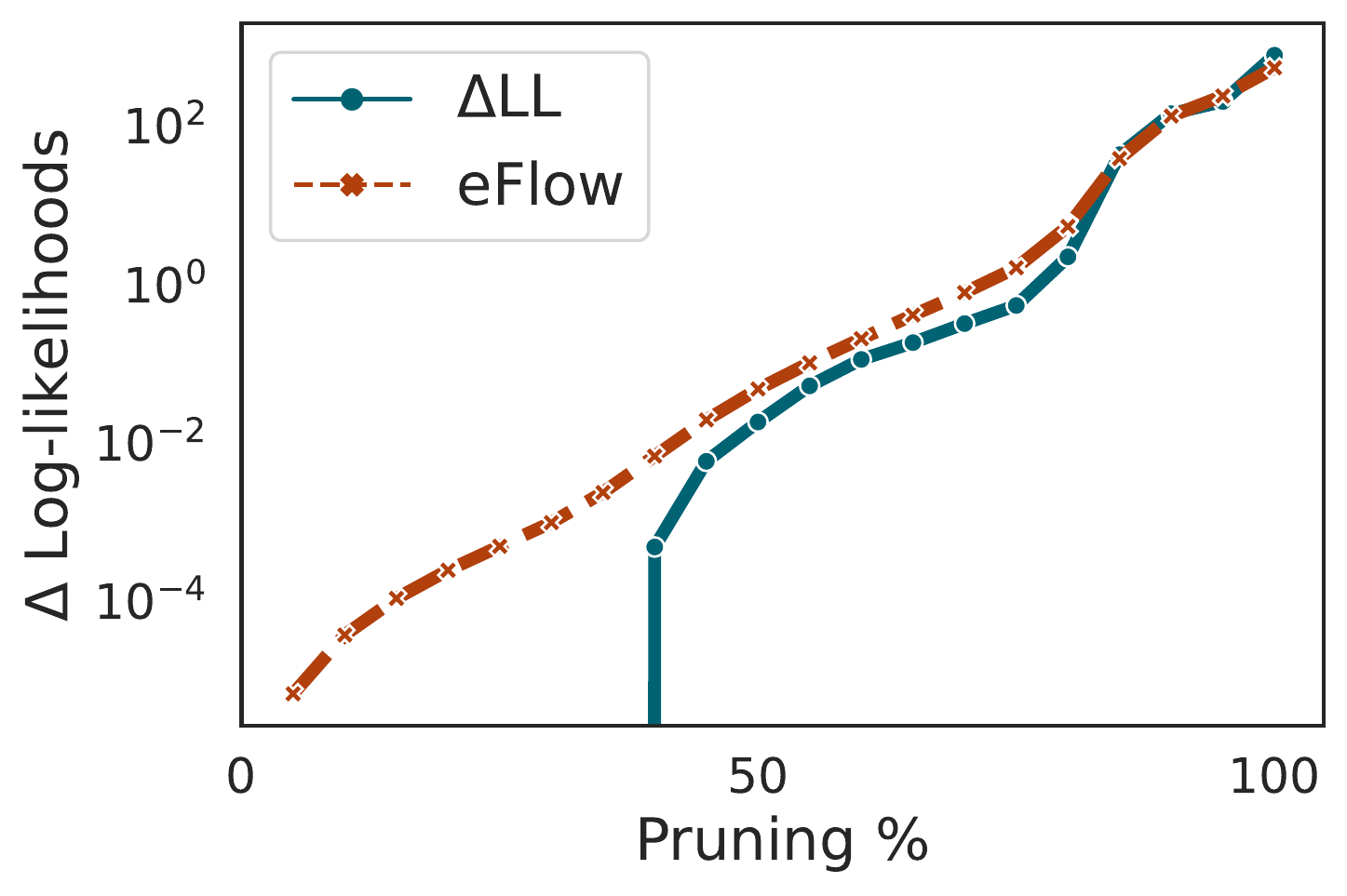}
    \caption{\label{fig:verify_heuristics} Comparing the actual loglikelihood drop ($\Delta$LL) and $\eflow$ heuristics (the approximated upper bound in Equation~\ref{eq:prune_mul_d_ll}). The approximate bound matches closely to the actual loglikelihood drop.}
    
\end{wrapfigure} Proof of this theorem is provided in Appendix~\ref{app-sec:proof-multi}. We first look at the second term of Equation~\ref{eq:prune_mul_d_ll}. Although it provides an upper bound to the performance drop, it cannot be used as a pruning heuristic since the bound does not decompose over edges. And hence finding the set of edges with the lowest score requires evaluating the bound exponentially many times with respect to the number of pruned edges.
Therefore, we do an additional approximation step of the bound via Taylor expansion, which leads to the third term of Equation~\ref{eq:prune_mul_d_ll}. This approximation matches the \eflow\ heuristic by a constant factor $1/|\data|$, which theoretically justifies the effectiveness of the heuristic. Figure~\ref{fig:verify_heuristics} empirically compares the actual log-likelihood drop and the quantity computed from the circuit flow heuristic (that is, the approximate upper bound) for different percentages of pruned parameters. We see that the approximate bound matches closely to the actual log-likelihood drop.

\section{Scalable Structure Learning}
\label{sec:struct-learn}

The pruning operator improves two aspects of PCs. First, as shown in Figure~\ref{fig:params_hist_after_prune}, model parameters are more balanced after pruning. Second, pruning removes sub-circuits with negligible contributions to the model's distribution. 
If we treat PCs as hierarchical mixtures of components, pruning can be regarded as an implicit structure learning step that removes the ``unimportant'' components for each mixture. 
However, since pruning only decreases model capacity, it is impossible to get a more expressive PC than the original one.
To mitigate this problem, we propose a \emph{growing} operation to increase the capacity of a PC by introducing more components for each mixture. 
Pruning and growing together define a scalable structure learning algorithm for PCs.

\begin{wrapfigure}{r}{0.40\linewidth}
    \centering
    \includegraphics[width=0.95\linewidth]{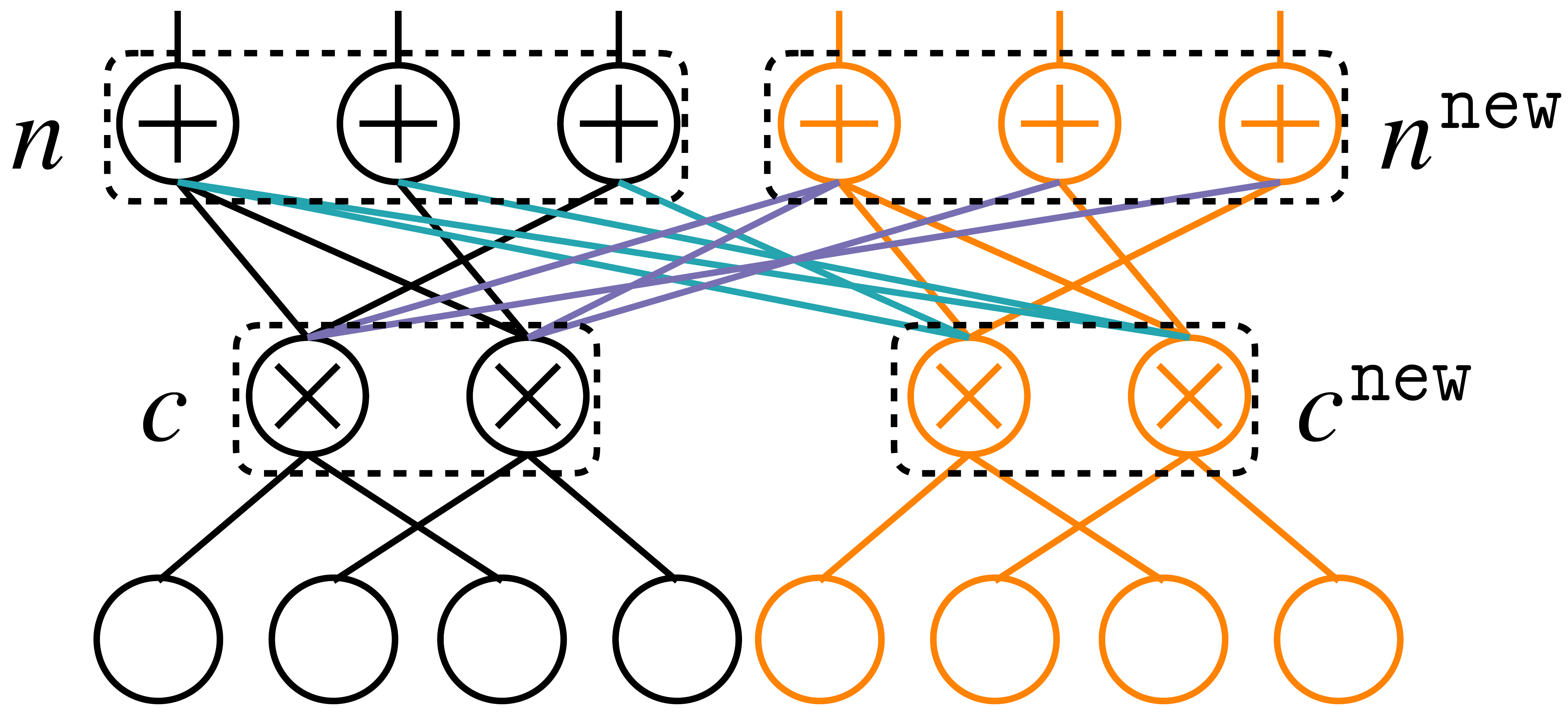}
    \caption{Growing operation. Each unit is doubled, and each parameterized edge is copied 3 times: 
    $(n^\new,c^\new)$ (\textcolor{gold2}{orange}), $(n^\new,c)$ (\textcolor{violet}{purple}), and $(n,c^\new)$ (\textcolor{petroil2}{green}).
    }
    \label{fig:op3}
\end{wrapfigure}
\paragraph{Growing.}
\emph{Growing} is an operator that increases model size by copying its existing components and injecting noise. As shown in Figure~\ref{fig:op}, after applying the growing operation on the original PC in Figure~\ref{fig:op2}, we can get a new grown PC as in Figure~\ref{fig:op3}. Specifically, the growing operation is applied to units, edges, and parameters respectively: (1) for units, growing operates on every PC unit $n$ and creates another copy $n^\new$; (2) for edges, the sum edge $(n,c)$ from the original PC (Figure~\ref{fig:op2}) are copied three times to the grown PC (Figure~\ref{fig:op3}): from new parent to new child $(n^\new,c^\new)$, from old parent to new child $(n,c^\new)$, and from new parent to old child $(n^\new,c)$; product edges are added to connect the copied version of a product unit and its copied inputs;
(3) a new parameter $\theta_{c\given n}^\new$ is a noisy copy of an old parameter $\theta_{c\given n}$, that is $\theta_{c\given n}^\new\leftarrow\epsilon\cdot\theta_{c\given n}$ where $\epsilon \sim \mathcal{N}(1, \sigma^2)$ and $\sigma^2$ controls the Gaussian noise variance. 
Gaussian noise is added to the copied parameters to ensure that after we apply the growing operation, parameter learning algorithms can find diverse parameters for different copies.
After a growing operation, the PC size is 4 times the original PC size.
Algorithm~\ref{alg:grow} in appendix shows a feedforward implementation of the growing operation.

\paragraph{Structure Learning through Pruning and Growing.}
The proposed pruning and growing algorithms can be applied iteratively to refine the structure and parameters of an initial PC. Specifically, since the growing operator increases the number of PC parameters by a factor of 4, applying growing after pruning 75\% of the edges from an initial PC keeps the number of parameters unchanged. 
We propose a joint structure and parameter learning algorithm for PCs that uses these two operations. 
Specifically, starting from an initial PC, we apply 75\% pruning, growing, and parameter learning iteratively until convergence. We utilize HCLTs~\citep{LiuNeurIPS21} as initial PC structure as it has the state-of-the-art likelihood performance. Note that this structure learning pipeline can be applied to any PC structure.

\paragraph{Parameter Estimation.}
We use a stochastic mini-batch version of Expectation-Maximization optimization~\citep{ChoiAAAI20}. Specifically, at each iteration, we draw a mini-batch of samples $\data_{B}$, compute aggregated circuit flows $\flow_{n,c}(\data_{B})$ and $\flow_{n}(\data_{B})$ of these samples (E-step), and then compute new parameter $\theta_{c\given n}^\new=\flow_{n,c}(\data_{B})/\flow_{n}(\data_{B})$. The parameters are then updated with learning rate $\alpha$: $\theta^{t+1} \leftarrow \alpha\theta^\new + (1-
\alpha)\theta^{t}$ (M-step). Empirically this approach converges faster and is better regularized compared to full-batch EM.

\paragraph{Parallel Computation.} Existing approaches to scaling up learning and inference with PCs, such as Einsum networks~\citep{peharz2020einsum}, utilize fully connected parametrized layers~(Figure~\ref{fig:op1}) of PC structures such as HCLT~\citep{LiuNeurIPS21} and RatSPN~\citep{peharz2020random}. These structures can be easily vectorized to utilize deep learning packages such as PyTorch. 
However, the sparse structure learned by pruning and growing is not easily vectorized as a dense matrix operation. We therefore implement customized GPU kernels to parallelize the computation of parameter learning and inference based on Juice.jl~\citep{DangAAAI21}, an open-source Julia package for learning PCs. The kernels segment PC units into layers such that the units in each layer are independent. Thus, the computation can be fully parallelized on the GPU. As a result, we can train PCs with millions of parameters in less than half an hour.

\section{Experiments}
\label{sec:exp}
We now evaluate our proposed method pruning and growing on two different sets of density estimation benchmarks: (1) the MNIST-family image generation datasets including MNIST~\citep{lecun2010mnist}, EMNIST~\citep{cohen2017emnist}, and FashionMNIST~\citep{xiao2017fashionmnist}; (2) the character-level Penn Tree Bank language modeling task~\citep{marcus19ptb}. 

Section~\ref{sec:exp-benchmark} first reports the best results we get on image datasets and language modeling tasks via the structure learning procedure proposed in Section~\ref{sec:struct-learn}. 
Section~\ref{sec:exp-ablate} then shows the effect of pruning and growing operations via two detailed experimental settings. It studies two different constrained optimization problems: finding the smallest PC for a given likelihood via model compression and finding the best PC of a given size via structure learning.

\paragraph{Settings.}For all experiments, we use hidden Chow-Liu Trees~(HCLTs)~\citep{LiuNeurIPS21} with the number of latent states in $\{16, 32, 64, 128\}$ as initial PC structures. We train the parameters of PCs with stochastic mini-batch EM (cf.~Section~\ref{sec:struct-learn}). We perform early stopping and hyperparameter search using a validation set and report results on the test set. Please refer to Appendix~\ref{app-sec:exp} for more details.
We use mean test set bits-per-dimension~(bpd) as the evaluation criteria, where
$\mathsf{bpd}(\data,\PC)=-\mathcal{LL}(\data,\PC)/(\log(2)  \cdot m)$ and $m$ is the number of features in dataset $\data$.

\subsection{Density Estimation Benchmarks}
\label{sec:exp-benchmark}
\paragraph{Image Datasets.}

The MNIST-family datasets contain gray-scale pixel images of size $28\times28$ where each pixel takes values in $[0,255]$. We split out 5\% of training data as a validation set. We compare with two competitive PC learning algorithms: HCLT~\citep{LiuNeurIPS21} and RatSPN~\citep{peharz2020random}, one flow-based model: IDF~\citep{hoogeboom2019integer}, and three VAE-based methods: BitSwap~\citep{kingma2019bit}, BB-ANS~\citep{townsend2018practical}, and McBits~\citep{ruan2021improving}. For a fair comparison, we implement RatSPN structures ourselves and use the same training pipeline and EM optimizer as our proposed method. Note that EinsumNet~\citep{peharz2020einsum} also uses RatSPN structures but with a PyTorch implementation so its comparison is subsumed by comparison with RatSPN. All 7 methods are tested on MNIST, 4 splits of EMNIST and FashionMNIST. As shown in Table~\ref{tab:mnist}, the best results are bold. We see that our proposed method significantly outperforms all other baselines on all datasets, and establishes new state-of-the-art results among PCs, flows, and VAE models. 
More experiment details are in Appendix~\ref{app-sec:exp}.
\begin{table}[ht]
    \footnotesize
    \caption{Density estimation performance on MNIST-family datasets in test set bpd.
    }
    \centering
    \centering
\linespread{1}

    \begin{tabular}{l||ccc|cccc}\toprule
Dataset         & Sparse PC (ours)         & HCLT  & RatSPN    & IDF   & BitSwap   & BB-ANS    & McBits\\\midrule
MNIST           & \textbf{1.14}     & 1.20  &   1.67   &1.90   &1.27       &1.39       &1.98   \\
EMNIST(MNIST)   & \textbf{1.52}              &  1.77  &  2.56   &  2.07  &   1.88    &   2.04   &   2.19 \\
EMNIST(Letters) & \textbf{1.58}     & 1.80  &   2.73  &1.95   &1.84       &2.26       &3.12   \\
EMNIST(Balanced)& \textbf{1.60}              & 1.82 &   2.78 &  2.15 & 1.96    &   2.23   &  2.88 \\
EMNIST(ByClass) & \textbf{1.54}     &1.85   &  2.72  &1.98   &1.87       &2.23       &3.14   \\
FashionMNIST    & \textbf{3.27}     & 3.34  &   4.29  &3.47   &3.28       &3.66       &3.72   \\\bottomrule
    \end{tabular}
    \label{tab:mnist}
\end{table}
\paragraph{Language Modeling Task.}

We use the Penn Tree Bank dataset with standard processing from~\citet{mikolov2012subword}, which contains around 5M characters and a character-level vocabulary size of $50$. The data is split into sentences with a maximum sequence length of $288$.
We compare with three competitive normalizing-flow-based models: Bipartite flow~\citep{tran2019discrete} and latent flows~\citep{ziegler2019latent} including AF/SCF and IAF/SCF, since they are the only comparable work with non-autoregressive language modeling.
As shown in Table~\ref{tab:ptbchar}, the proposed method outperforms all three baselines.
\begin{table}[ht]
    \footnotesize
    \caption{\label{tab:ptbchar}Character-level language modeling results on Penn Tree Bank in test set bpd.}
    \centering
    \centering
\linespread{1}
    \begin{tabular}{l||c|ccc}\toprule
Dataset         & Sparse PC  (ours)          & Bipartite flow~\citep{tran2019discrete}    & AF/SCF~\citep{ziegler2019latent}    & IAF/SCF~\citep{ziegler2019latent} \\\midrule
Penn Tree Bank  & \textbf{1.35} & 1.38              &1.46       &1.63\\\bottomrule
    \end{tabular}
\end{table}

\subsection{Evaluating Pruning and Growing}
\label{sec:exp-ablate}

\begin{figure}[ht]
    \centering
    \includegraphics[width=0.175\textwidth]{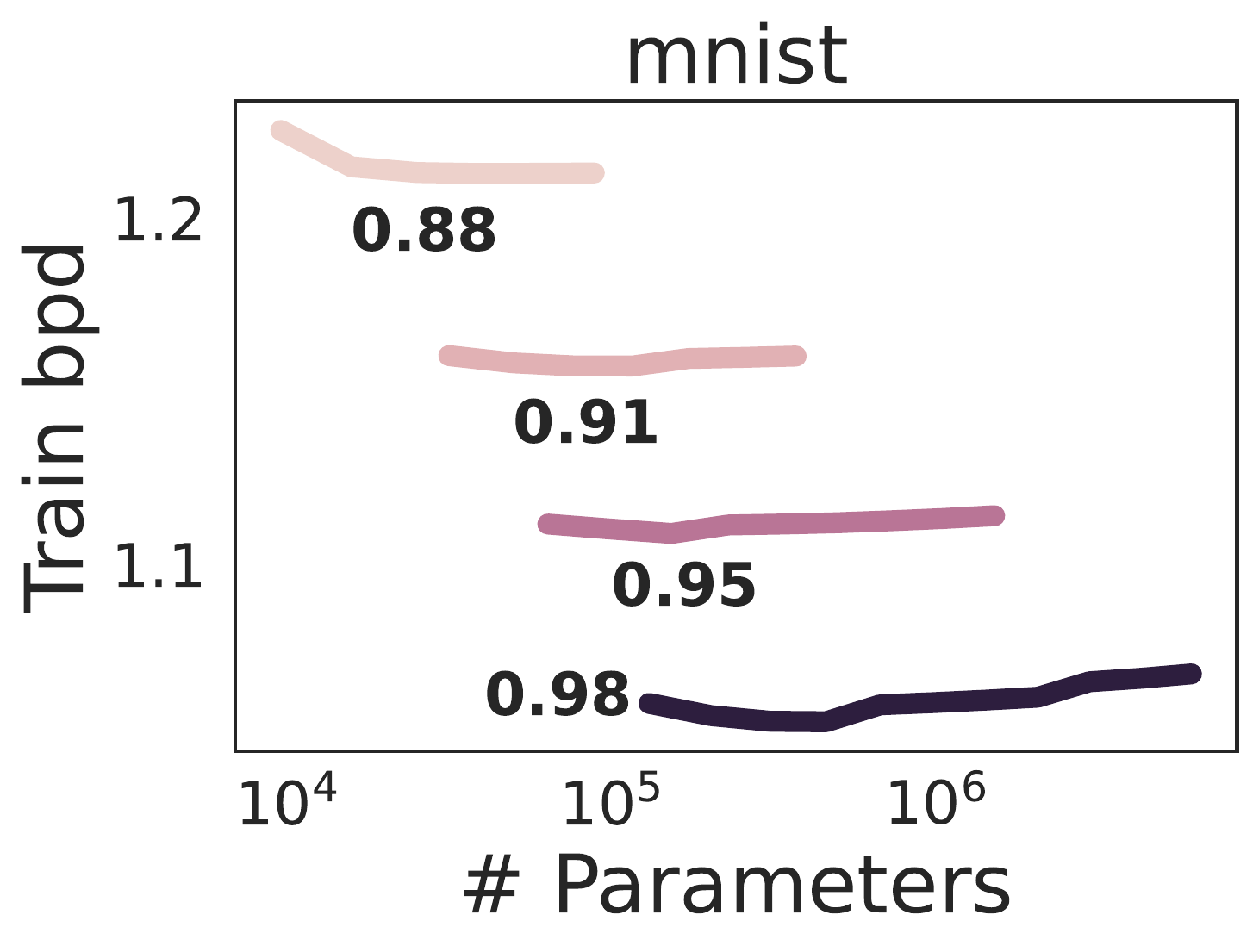}
    \includegraphics[width=0.157\textwidth]{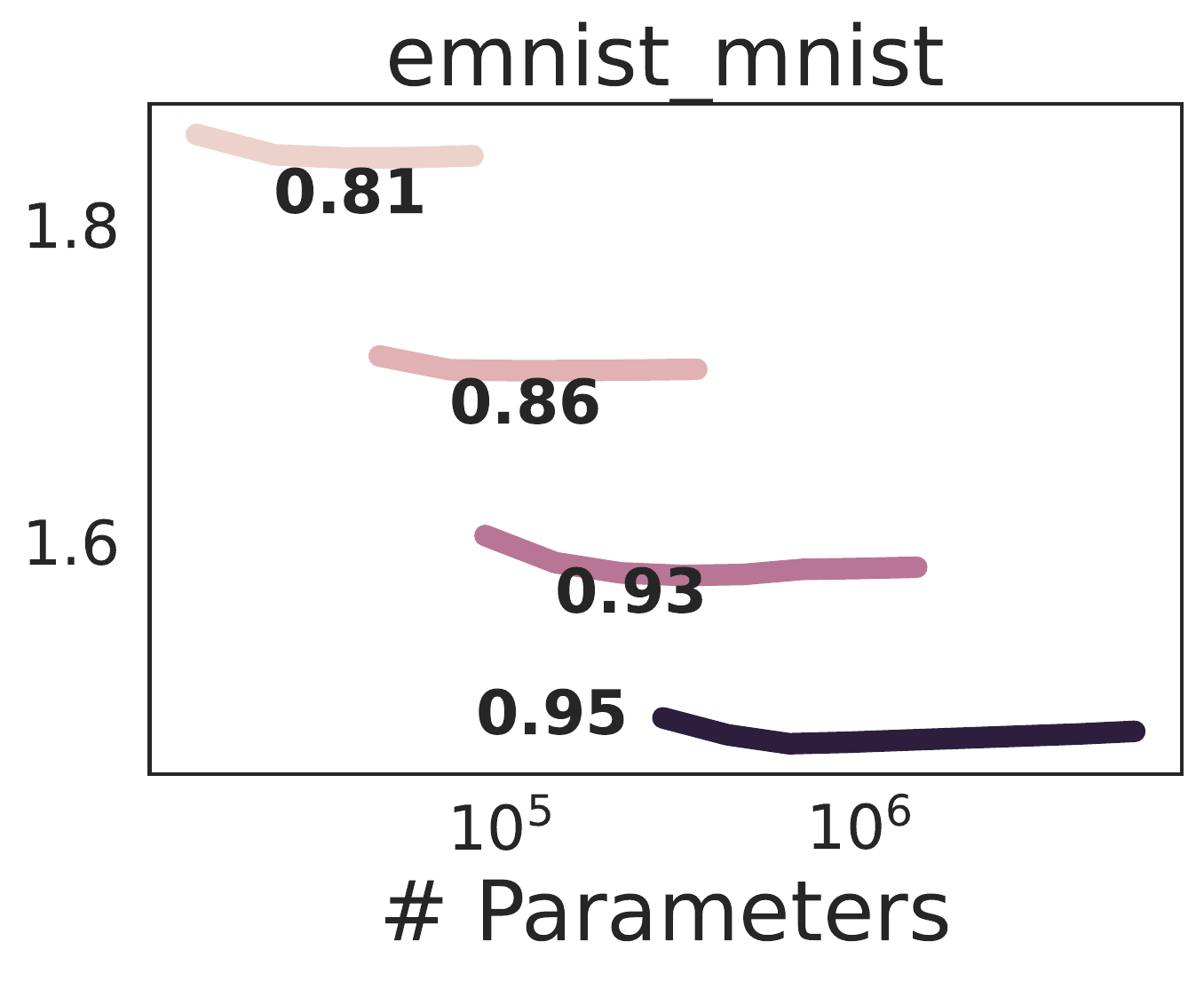}
    \includegraphics[width=0.157\textwidth]{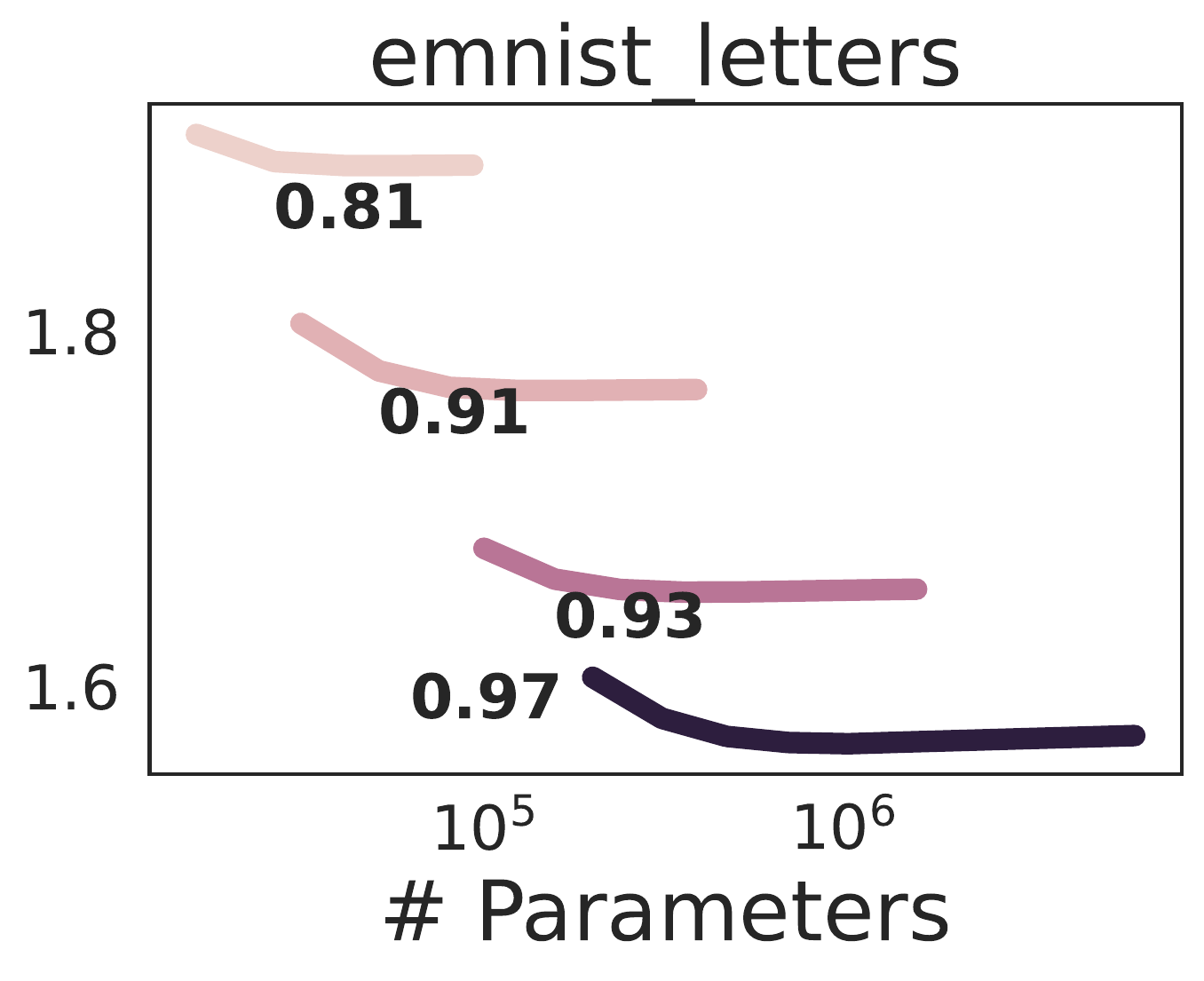}
    \includegraphics[width=0.157\textwidth]{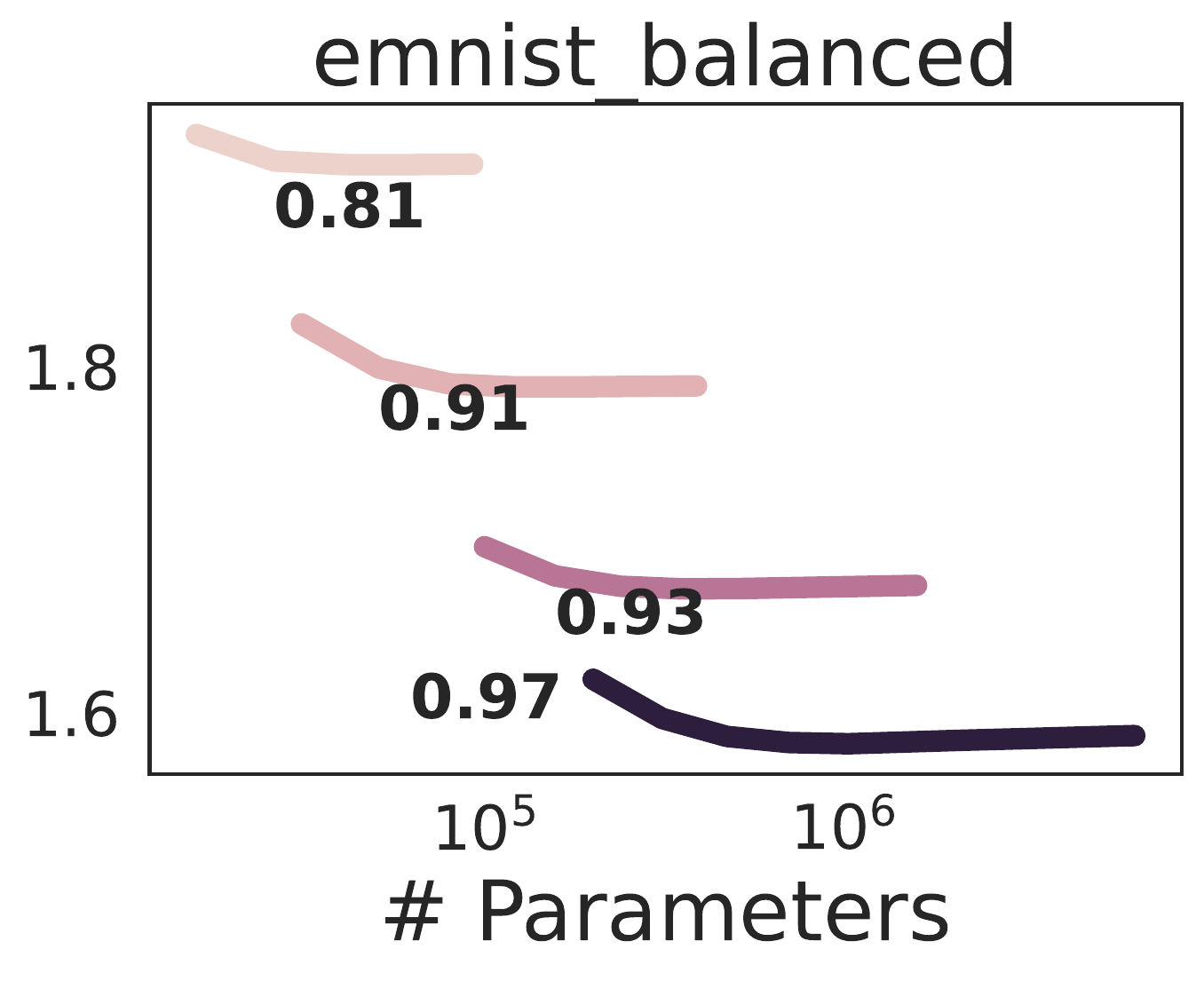}
    \includegraphics[width=0.157\textwidth]{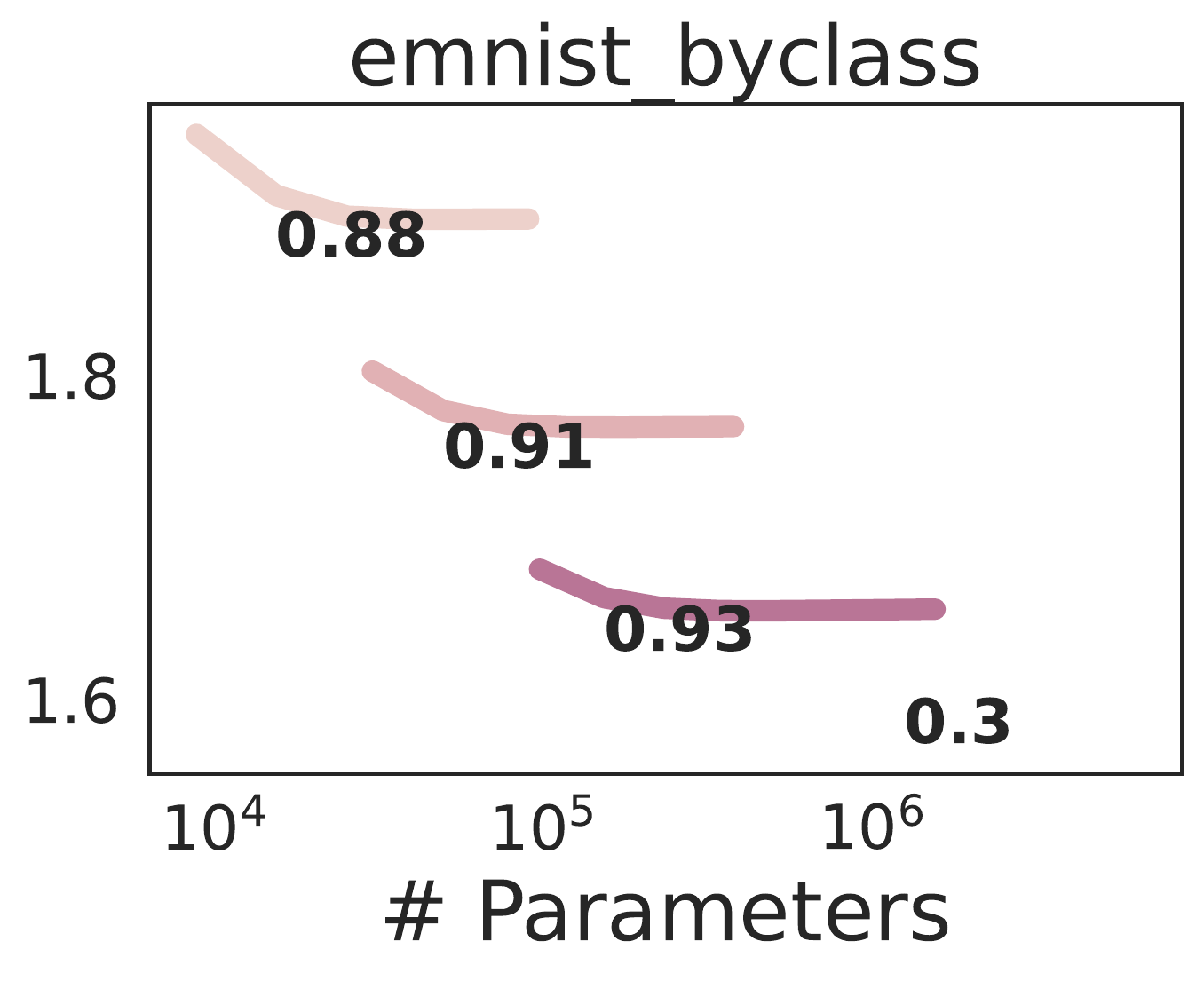}
    \includegraphics[width=0.158\textwidth]{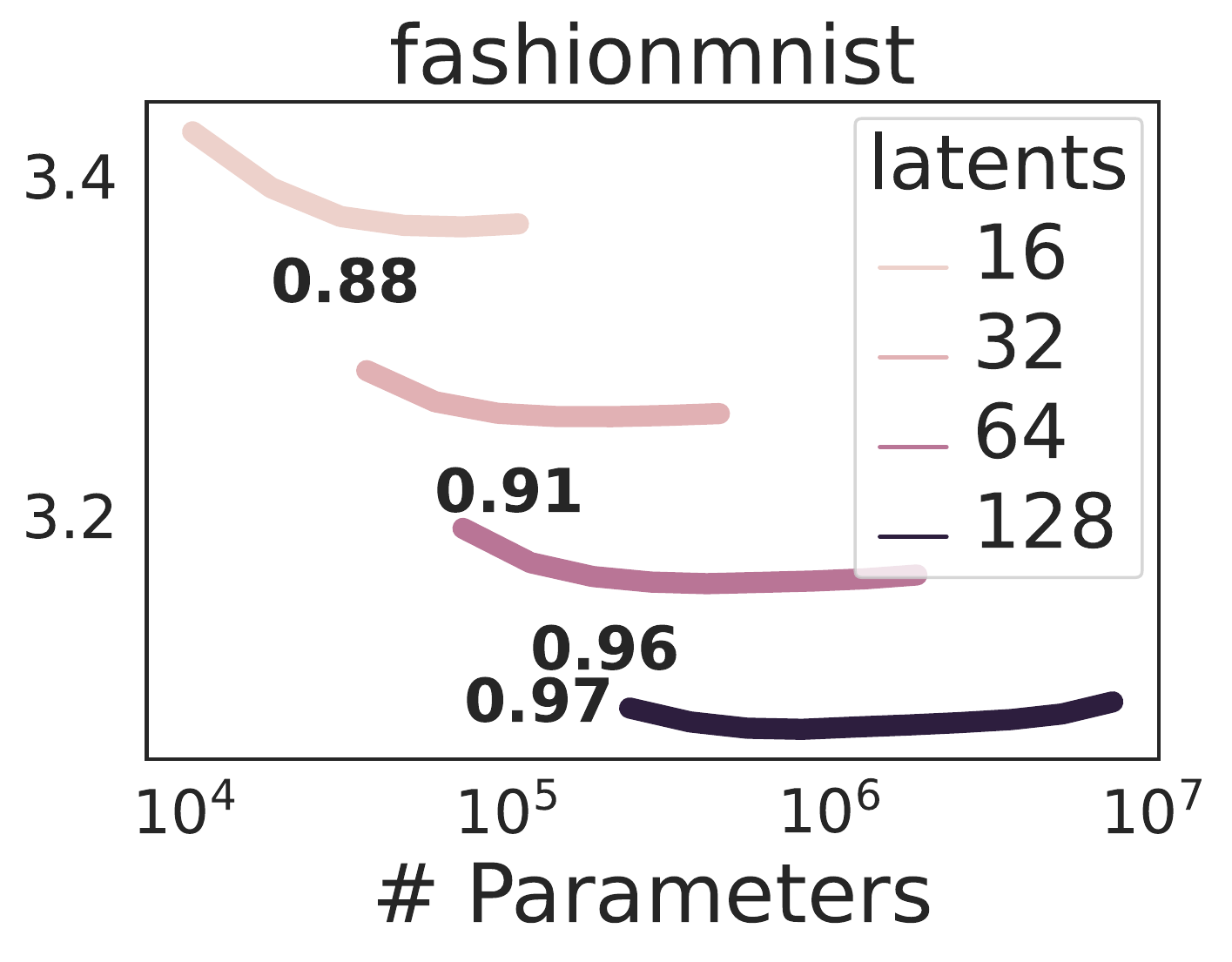}
    \caption{\label{fig:distll}Model compression via pruning and finetuning. We report the training set bpd (y-axis) in terms of the number of parameters (x-axis) for different numbers of latent states. For each curve, compression starts from the right (initial PC \#Params $|\PC^{\mathtt{init}}|$) and ends at the left (compressed PC \#Params $|\PC^{\mathtt{com}}|$); compression rate (1 - $|\PC^{\mathtt{com}}|$ / $|\PC^{\mathtt{init}}|$) is annotated next to each curve. }
\end{figure}

\paragraph{What is the Smallest PC for the Same Likelihood?}
We evaluate the ability of pruning based on circuit flows to do effective model compression by iteratively pruning a $k$-fraction of the PC parameters and then fine-tuning them until the final training log-likelihood does not decrease by more than $1\%$. Specifically, we take pruning percentage $k$ from $\{0.05, 0.1, 0.3\}$.
As shown in Figure~\ref{fig:distll}, we can achieve a compression rate of 80-98\% with negligible performance loss on PCs. Besides, by fixing the number of latent parameters (x-axis) and comparing bpp across different numbers of latent states (legend), we discover that compressing a large PC to a get smaller PC yields better likelihoods compared to directly training an HCLT with the same number of parameters from scratch. This can be explained by the sparsity of compressed PC structures, as well as a smarter way of finding good parameters: learning a better PC with larger size and compressing it down to a smaller one.

\begin{figure}[ht]
    \centering
    \includegraphics[width=0.17\textwidth]{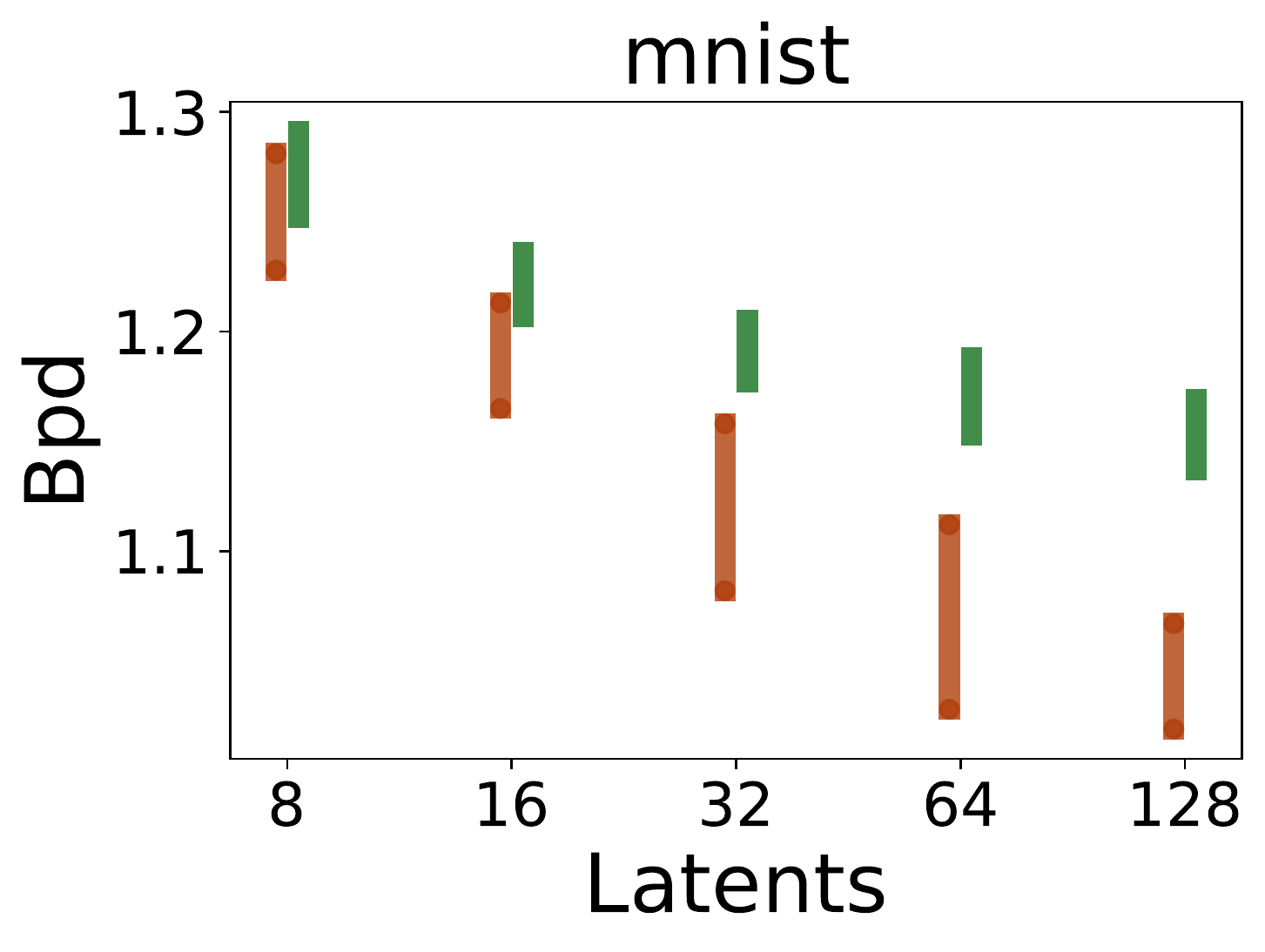}
    \includegraphics[width=0.16\textwidth]{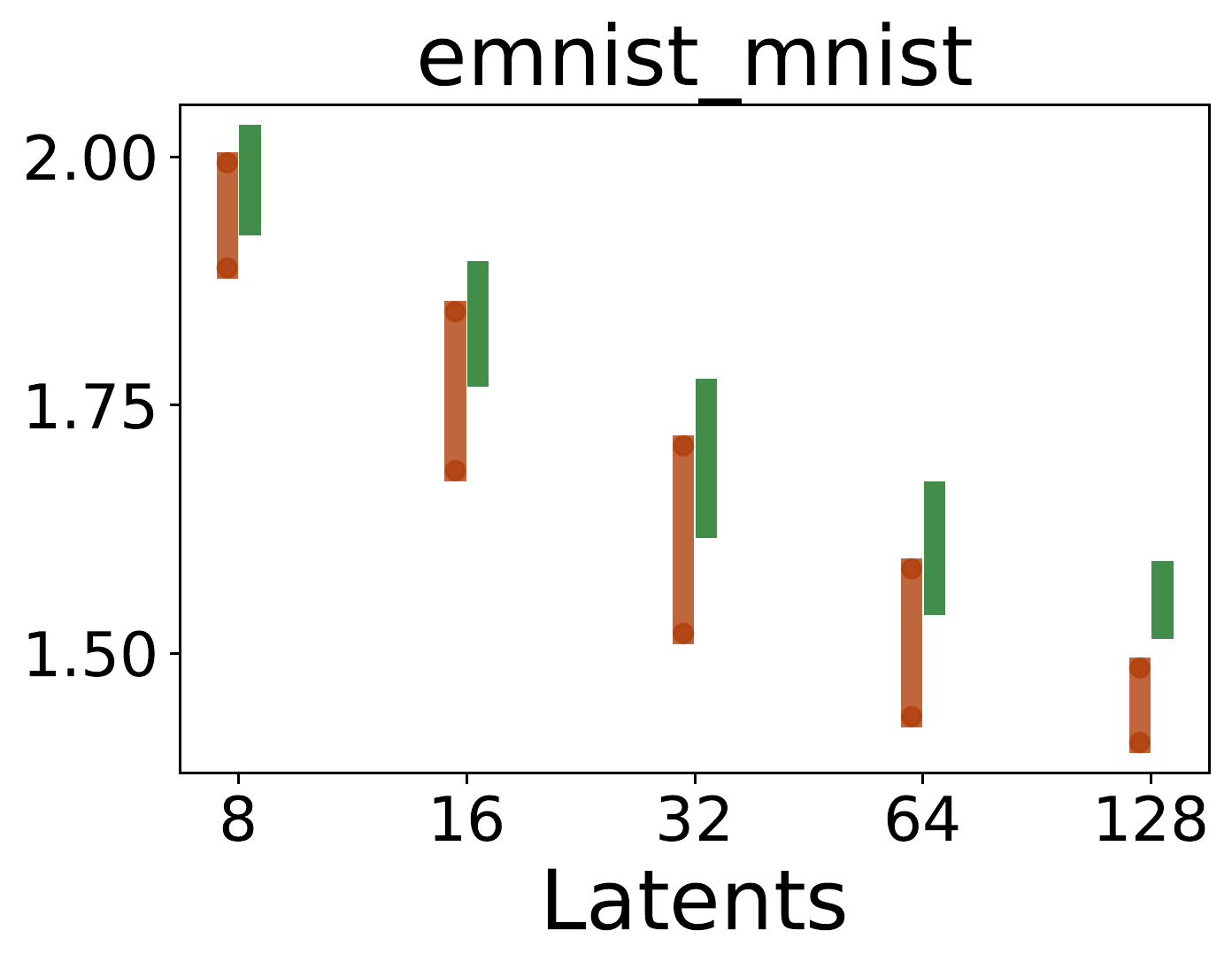}
    \includegraphics[width=0.16\textwidth]{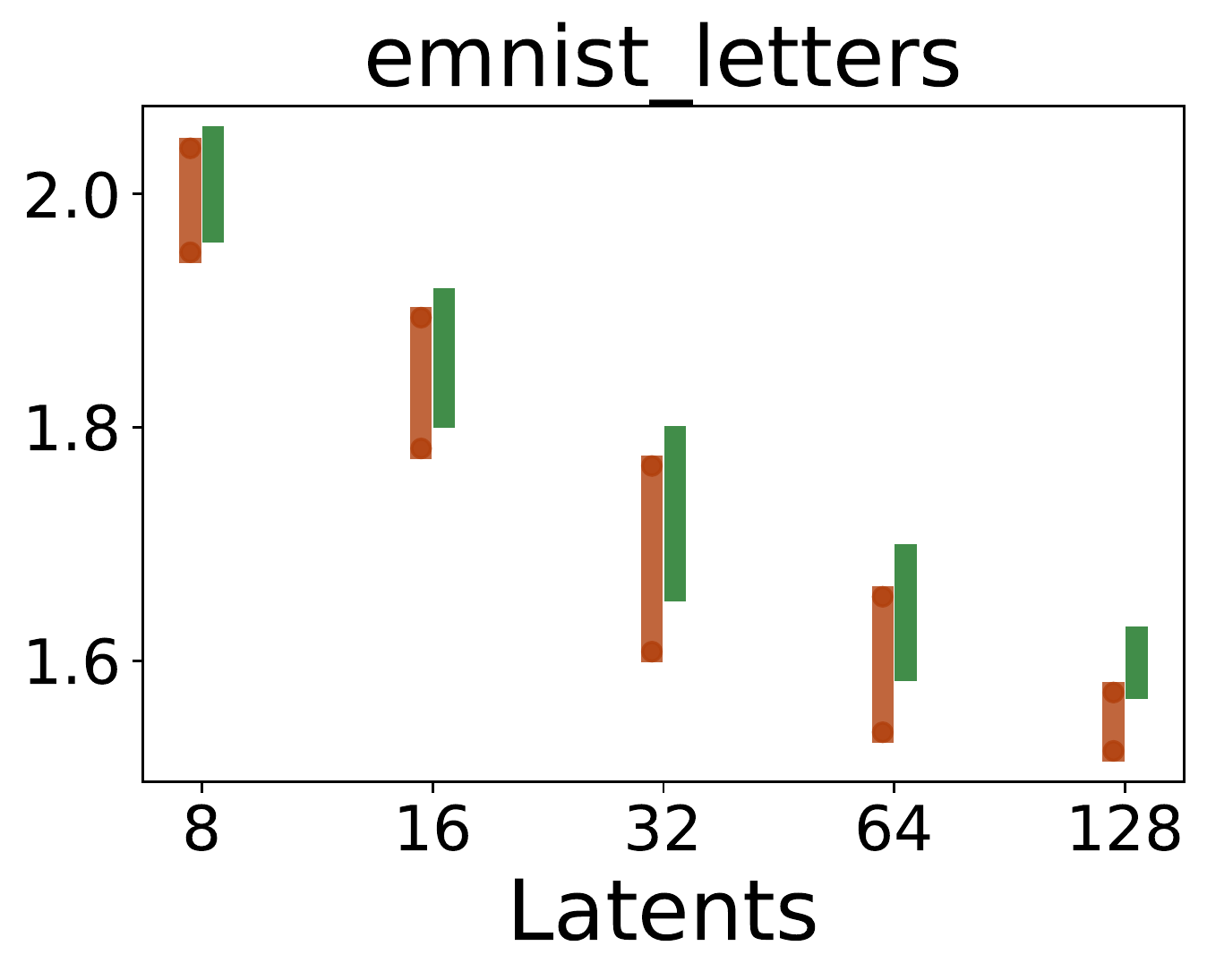}
    \includegraphics[width=0.16\textwidth]{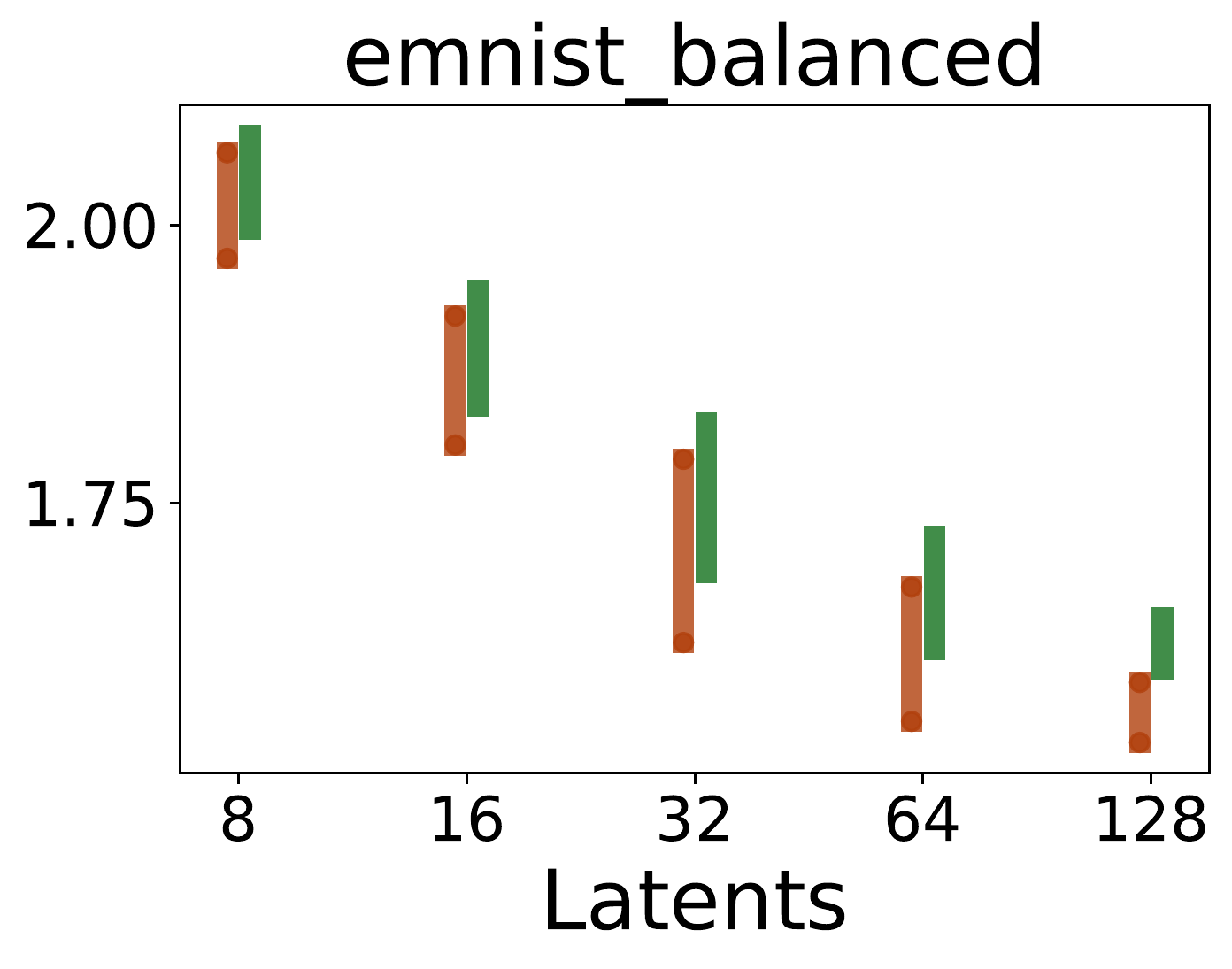}
    \includegraphics[width=0.16\textwidth]{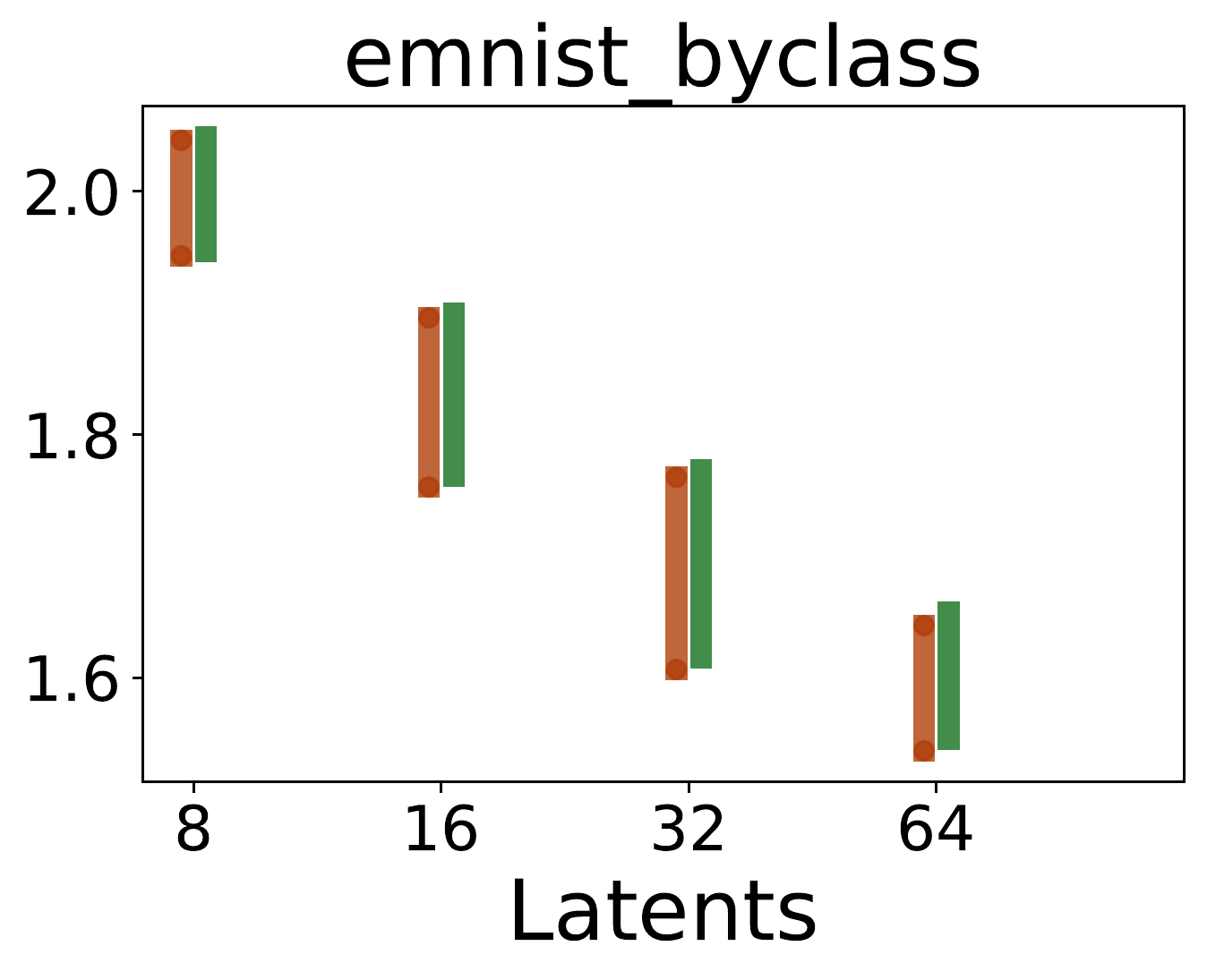}
    \includegraphics[width=0.16\textwidth]{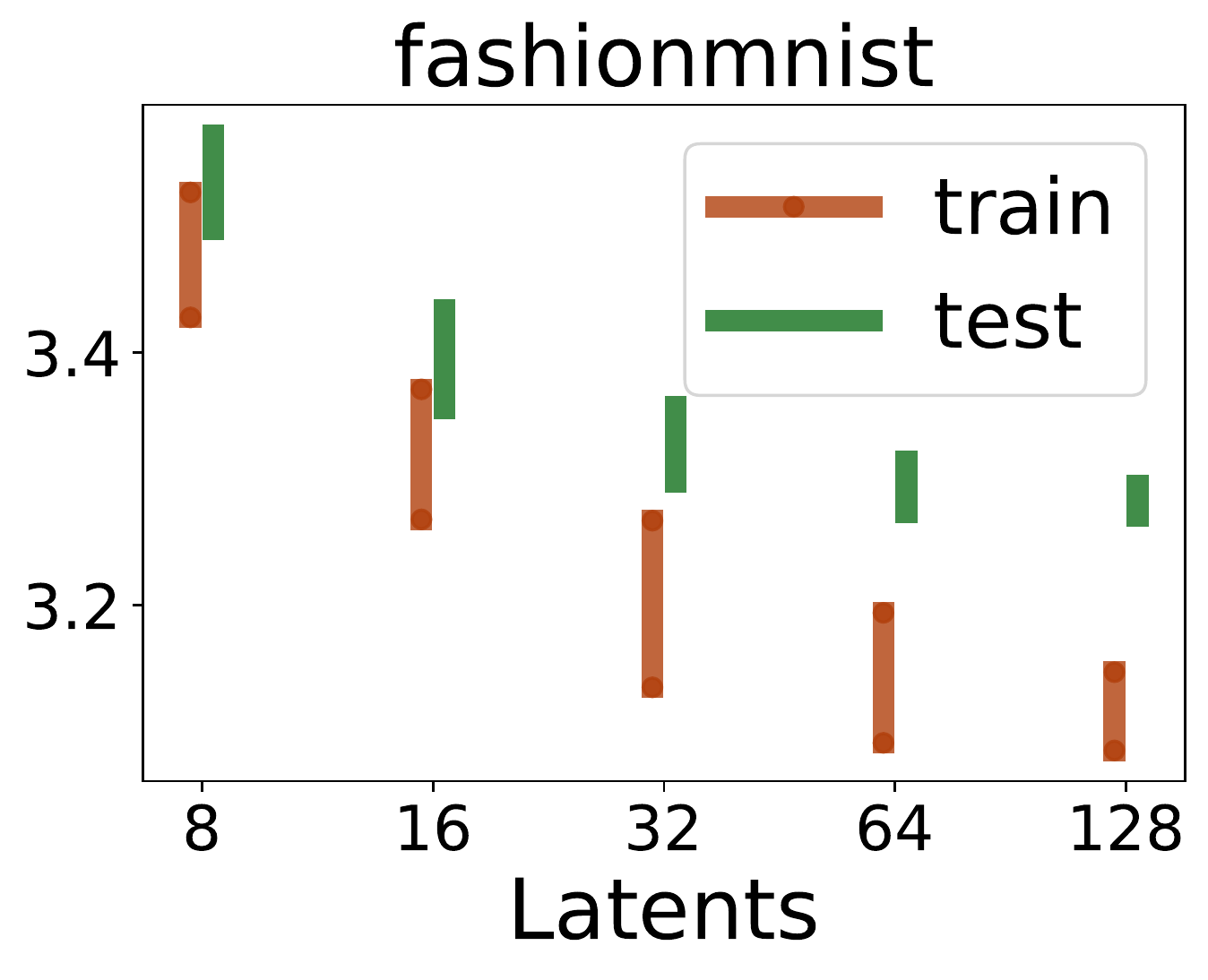}
    \caption{\label{fig:samesize}Structure learning via 75\% pruning, growing and finetuning. We report bpd~(y-axis) on both train (red) and test set (green) in terms of the number of latent states (x-axis). For each curve, training starts from the top (large bpd) and ends at the bottom (small bpd).}
\end{figure}

\paragraph{What is the Best PC for the Same Size?}
We evaluate structure learning that combines pruning and growing as proposed in Section~\ref{sec:struct-learn}. Starting from an initial HCLT, we iteratively prune 75\% of the parameters, grow again, and fine-tune until meeting the stopping criteria. As shown in Figure~\ref{fig:samesize}, our method consistently improve the likelihoods of initial PCs for different numbers of latent states among all datasets.

\section{Related Work}
Improving the expressiveness of PCs has been a central topic in the literature. Predominant works focus on the structure learning algorithms that iteratively modify PC structures to progressively fit the data \citep{DangPGM20,LiangUAI17,robert13learnspn}. Alternatively, a recent trend is to construct PCs with good initial structures 
and only perform parameter learning afterwards \citep{rahman2014cutset, Adel15svd,ZhangICML21}. For example, EiNets \citep{peharz2020einsum}, RAT-SPNs \citep{peharz2020random}, and XPCs \citep{Mauro21random}  use randomly generated structures, HCLTs \citep{LiuNeurIPS21} and ID-SPNs~\citep{rooshenas2014learning} define PCs dependent on the pairwise correlation on variables, and learn direct and indirect variable correlations among variables. 
There are also a few works that boost PC performance with the expressive power of neural networks. CSPNs \citep{shao2022conditional} harness neural networks to learn expressive conditional density estimators, and HyperSPN \citep{shih2021hyperspns} utilizes neural networks for better PC regularization. 

Pruning and growing have been introduced in deep neural networks to exploit sparsity \citep{Torsten21}.
Similar strategies can be adopted in probabilistic circuits. \citet{patil2010evaluation} prune decision trees according to validation accuracy reduction. ResSPNs uses the lottery ticket hypothesis for weight pruning to gain more compact PCs \citep{ventola2020residual}. However, the neural network pruning methods and above PC pruning methods mainly focus on pruning by parameters. In contrast to these, our work develops better pruning strategies based on semantic properties of PCs.

\section{Conclusions}
\label{sec:conclusion}
We propose structure learning of probabilistic circuits by combining pruning and growing operations to exploit the sparsity of PC structures. We show significant empirical improvements in the density estimation tasks of PCs compared to existing PC learners and competing flow-based models and~VAEs. 
All our Sparse-PC learning and inference algorithms are available as GPU-parallel implementations.

\paragraph{Acknowledgements}
This work was funded in part by the DARPA Perceptually-enabled Task Guidance (PTG) Program under contract number HR00112220005, NSF grants \#IIS-1943641, \#IIS-1956441, \#CCF-1837129, Samsung, CISCO, a Sloan Fellowship, and a UCLA Samueli Fellowship. We thank Honghua Zhang for proofreading and insightful comments on this paper's final version.
\bibliographystyle{plainnat}
\bibliography{references}

\begin{thebibliography}{48}
\providecommand{\natexlab}[1]{#1}
\providecommand{\url}[1]{\texttt{#1}}
\expandafter\ifx\csname urlstyle\endcsname\relax
  \providecommand{\doi}[1]{doi: #1}\else
  \providecommand{\doi}{doi: \begingroup \urlstyle{rm}\Url}\fi

\bibitem[Adel et~al.(2015)Adel, Balduzzi, and Ghodsi]{Adel15svd}
Tameem Adel, David Balduzzi, and Ali Ghodsi.
\newblock Learning the structure of sum-product networks via an svd-based
  algorithm.
\newblock In \emph{Proceedings of the 31st Conference on Uncertainty in
  Artificial Intelligence (UAI)}, 2015.

\bibitem[Choi et~al.(2020{\natexlab{a}})Choi, Farnadi, Babaki, and Van~den
  Broeck]{ChoiAAAI20}
YooJung Choi, Golnoosh Farnadi, Behrouz Babaki, and Guy Van~den Broeck.
\newblock Learning fair naive bayes classifiers by discovering and eliminating
  discrimination patterns.
\newblock In \emph{Proceedings of the 34th AAAI Conference on Artificial
  Intelligence}, 2020{\natexlab{a}}.

\bibitem[Choi et~al.(2020{\natexlab{b}})Choi, Vergari, and Van~den
  Broeck]{ProbCirc20}
YooJung Choi, Antonio Vergari, and Guy Van~den Broeck.
\newblock Probabilistic circuits: A unifying framework for tractable
  probabilistic models.
\newblock \emph{Technical report}, 2020{\natexlab{b}}.

\bibitem[Choi et~al.(2021)Choi, Dang, and Van~den Broeck]{ChoiAAAI21}
YooJung Choi, Meihua Dang, and Guy Van~den Broeck.
\newblock Group fairness by probabilistic modeling with latent fair decisions.
\newblock In \emph{Proceedings of the 35th AAAI Conference on Artificial
  Intelligence}, 2021.

\bibitem[Cohen et~al.(2017)Cohen, Afshar, Tapson, and
  Van~Schaik]{cohen2017emnist}
Gregory Cohen, Saeed Afshar, Jonathan Tapson, and Andre Van~Schaik.
\newblock Emnist: Extending mnist to handwritten letters.
\newblock In \emph{2017 International Joint Conference on Neural Networks
  (IJCNN)}, 2017.

\bibitem[Correia et~al.(2020)Correia, Peharz, and de~Campos]{correia2020joints}
Alvaro Correia, Robert Peharz, and Cassio~P de~Campos.
\newblock Joints in random forests.
\newblock In \emph{Advances in Neural Information Processing Systems 33
  (NeurIPS)}, 2020.

\bibitem[Dang et~al.(2020)Dang, Vergari, and Van~den Broeck]{DangPGM20}
Meihua Dang, Antonio Vergari, and Guy Van~den Broeck.
\newblock Strudel: Learning structured-decomposable probabilistic circuits.
\newblock In \emph{Proceedings of the 10th International Conference on
  Probabilistic Graphical Models (PGM)}, 2020.

\bibitem[Dang et~al.(2021)Dang, Khosravi, Liang, Vergari, and Van~den
  Broeck]{DangAAAI21}
Meihua Dang, Pasha Khosravi, Yitao Liang, Antonio Vergari, and Guy Van~den
  Broeck.
\newblock Juice: A julia package for logic and probabilistic circuits.
\newblock In \emph{Proceedings of the 35th AAAI Conference on Artificial
  Intelligence (Demo Track)}, 2021.

\bibitem[Darwiche(2002)]{darwiche02KR}
Adnan Darwiche.
\newblock A logical approach to factoring belief networks.
\newblock In \emph{Proceedings of the 8th International Conference on
  Principles of Knowledge Representation and Reasoning (KR)}, 2002.

\bibitem[Darwiche(2003)]{darwicheJACM-POLY}
Adnan Darwiche.
\newblock A differential approach to inference in bayesian networks.
\newblock \emph{Journal of the ACM}, 2003.

\bibitem[Darwiche and Marquis(2002)]{darwiche2002knowledge}
Adnan Darwiche and Pierre Marquis.
\newblock A knowledge compilation map.
\newblock \emph{Journal of Artificial Intelligence Research}, 2002.

\bibitem[Di~Mauro et~al.(2021)Di~Mauro, Gala, Iannotta, and
  Basile]{Mauro21random}
Nicola Di~Mauro, Gennaro Gala, Marco Iannotta, and Teresa~M.A. Basile.
\newblock Random probabilistic circuits.
\newblock In \emph{Proceedings of the 37th Conference on Uncertainty in
  Artificial Intelligence (UAI)}, 2021.

\bibitem[Dietrichstein et~al.(2022)Dietrichstein, Major, Trapp, Wimmer, Lenis,
  Winter, Berg, Neubauer, and B{\"u}hler]{dietrichstein2022anomaly}
Marc Dietrichstein, David Major, Martin Trapp, Maria Wimmer, Dimitrios Lenis,
  Philip Winter, Astrid Berg, Theresa Neubauer, and Katja B{\"u}hler.
\newblock Anomaly detection using generative models and sum-product networks in
  mammography scans.
\newblock In \emph{MICCAI Workshop on Deep Generative Models}, 2022.

\bibitem[Dinh et~al.(2014)Dinh, Krueger, and Bengio]{dinh2014nice}
Laurent Dinh, David Krueger, and Yoshua Bengio.
\newblock {Nice}: Non-linear independent components estimation.
\newblock \emph{arXiv preprint arXiv:1410.8516}, 2014.

\bibitem[Gens and Pedro(2013)]{robert13learnspn}
Robert Gens and Domingos Pedro.
\newblock Learning the structure of sum-product networks.
\newblock In \emph{Proceedings of the 30th International Conference on Machine
  Learning (ICML)}, 2013.

\bibitem[Hoefler et~al.(2021)Hoefler, Alistarh, Ben-Nun, Dryden, and
  Peste]{Torsten21}
Torsten Hoefler, Dan Alistarh, Tal Ben-Nun, Nikoli Dryden, and Alexandra Peste.
\newblock Sparsity in deep learning: Pruning and growth for efficient inference
  and training in neural networks.
\newblock \emph{Journal of Machine Learning Research}, 2021.

\bibitem[Hoogeboom et~al.(2019)Hoogeboom, Peters, Van Den~Berg, and
  Welling]{hoogeboom2019integer}
Emiel Hoogeboom, Jorn Peters, Rianne Van Den~Berg, and Max Welling.
\newblock Integer discrete flows and lossless compression.
\newblock In \emph{Advances in Neural Information Processing Systems 32
  (NeurIPS)}, 2019.

\bibitem[Khosravi et~al.(2019)Khosravi, Choi, Liang, Vergari, and Van~den
  Broeck]{khosravi2019tractable}
Pasha Khosravi, YooJung Choi, Yitao Liang, Antonio Vergari, and Guy Van~den
  Broeck.
\newblock On tractable computation of expected predictions.
\newblock In \emph{Advances in Neural Information Processing Systems 32
  (NeurIPS)}, 2019.

\bibitem[Kingma and Welling(2013)]{kingma2013auto}
Diederik~P Kingma and Max Welling.
\newblock Auto-encoding variational bayes.
\newblock \emph{arXiv preprint arXiv:1312.6114}, 2013.

\bibitem[Kingma et~al.(2019)Kingma, Abbeel, and Ho]{kingma2019bit}
Friso Kingma, Pieter Abbeel, and Jonathan Ho.
\newblock Bit-swap: Recursive bits-back coding for lossless compression with
  hierarchical latent variables.
\newblock In \emph{Proceedings of the 36th International Conference on Machine
  Learning (ICML)}, 2019.

\bibitem[Kisa et~al.(2014)Kisa, {Van den Broeck}, Choi, and
  Darwiche]{KisaVCD14}
Doga Kisa, Guy {Van den Broeck}, Arthur Choi, and Adnan Darwiche.
\newblock Probabilistic sentential decision diagrams.
\newblock In \emph{Proceedings of the 14th International Conference on
  Principles of Knowledge Representation and Reasoning (KR)}, 2014.

\bibitem[LeCun et~al.(2010)LeCun, Cortes, and Burges]{lecun2010mnist}
Yann LeCun, Corinna Cortes, and CJ~Burges.
\newblock Mnist handwritten digit database.
\newblock \emph{ATT Labs [Online]. Available:
  http://yann.lecun.com/exdb/mnist}, 2, 2010.

\bibitem[Li et~al.(2021)Li, Zeng, Vergari, and Van~den Broeck]{LiUAI21}
Wenzhe Li, Zhe Zeng, Antonio Vergari, and Guy Van~den Broeck.
\newblock Tractable computation of expected kernels.
\newblock In \emph{Proceedings of the 37th Conference on Uncertainty in
  Aritifical Intelligence (UAI)}, 2021.

\bibitem[Liang et~al.(2017)Liang, Bekker, and Van~den Broeck]{LiangUAI17}
Yitao Liang, Jessa Bekker, and Guy Van~den Broeck.
\newblock Learning the structure of probabilistic sentential decision diagrams.
\newblock In \emph{Proceedings of the 33rd Conference on Uncertainty in
  Artificial Intelligence (UAI)}, 2017.

\bibitem[Liu and Van~den Broeck(2021)]{LiuNeurIPS21}
Anji Liu and Guy Van~den Broeck.
\newblock Tractable regularization of probabilistic circuits.
\newblock In \emph{Advances in Neural Information Processing Systems 34
  (NeurIPS)}, 2021.

\bibitem[Liu et~al.(2022)Liu, Mandt, and Van~den Broeck]{LiuICLR22lossless}
Anji Liu, Stephan Mandt, and Guy Van~den Broeck.
\newblock Lossless compression with probabilistic circuits.
\newblock In \emph{Proceedings of the 10th International Conference on Learning
  Representations (ICLR)}, 2022.

\bibitem[Marcus et~al.(1993)Marcus, Marcinkiewicz, and Santorini]{marcus19ptb}
Mitchell~P. Marcus, Mary~Ann Marcinkiewicz, and Beatrice Santorini.
\newblock Building a large annotated corpus of english: The penn treebank.
\newblock \emph{Computational Linguistics}, 1993.

\bibitem[Marinescu and Dechter(2005)]{marinescu2005and}
Radu Marinescu and Rina Dechter.
\newblock And/or branch-and-bound for graphical models.
\newblock In \emph{Proceedings of the 19th International Joint Conference on
  Artificial Intelligence (IJCAI)}, 2005.

\bibitem[Mikolov et~al.(2012)Mikolov, Sutskever, Deoras, Le, and
  Kombrink]{mikolov2012subword}
Tom{\'a}{\v{s}} Mikolov, Ilya Sutskever, Anoop Deoras, Hai-Son Le, and Stefan
  Kombrink.
\newblock Subword language modeling with neural networks.
\newblock \emph{preprint (http://www.fit.vutbr.cz/imikolov/rnnlm/char.pdf)},
  2012.

\bibitem[Molina et~al.(2019)Molina, Vergari, Stelzner, Peharz, Subramani,
  Di~Mauro, Poupart, and Kersting]{molina2019spflow}
Alejandro Molina, Antonio Vergari, Karl Stelzner, Robert Peharz, Pranav
  Subramani, Nicola Di~Mauro, Pascal Poupart, and Kristian Kersting.
\newblock Spflow: An easy and extensible library for deep probabilistic
  learning using sum-product networks.
\newblock \emph{arXiv preprint arXiv:1901.03704}, 2019.

\bibitem[Papamakarios et~al.(2021)Papamakarios, Nalisnick, Rezende, Mohamed,
  and Lakshminarayanan]{papamakarios2021normalizing}
George Papamakarios, Eric Nalisnick, Danilo~Jimenez Rezende, Shakir Mohamed,
  and Balaji Lakshminarayanan.
\newblock Normalizing flows for probabilistic modeling and inference.
\newblock \emph{Journal of Machine Learning Research}, 2021.

\bibitem[Patil et~al.(2010)Patil, Wadhai, and Gokhale]{patil2010evaluation}
Dipti~D Patil, VM~Wadhai, and JA~Gokhale.
\newblock Evaluation of decision tree pruning algorithms for complexity and
  classification accuracy.
\newblock \emph{International Journal of Computer Applications}, 2010.

\bibitem[Peharz et~al.(2020{\natexlab{a}})Peharz, Lang, Vergari, Stelzner,
  Molina, Trapp, Van~den Broeck, Kersting, and Ghahramani]{peharz2020einsum}
Robert Peharz, Steven Lang, Antonio Vergari, Karl Stelzner, Alejandro Molina,
  Martin Trapp, Guy Van~den Broeck, Kristian Kersting, and Zoubin Ghahramani.
\newblock Einsum networks: Fast and scalable learning of tractable
  probabilistic circuits.
\newblock In \emph{Proceedings of the 37th International Conference on Machine
  Learning (ICML)}, 2020{\natexlab{a}}.

\bibitem[Peharz et~al.(2020{\natexlab{b}})Peharz, Vergari, Stelzner, Molina,
  Shao, Trapp, Kersting, and Ghahramani]{peharz2020random}
Robert Peharz, Antonio Vergari, Karl Stelzner, Alejandro Molina, Xiaoting Shao,
  Martin Trapp, Kristian Kersting, and Zoubin Ghahramani.
\newblock Random sum-product networks: A simple and effective approach to
  probabilistic deep learning.
\newblock In \emph{Proceedings of the 36th Conference on Uncertainty in
  Aritifical Intelligence (UAI)}, 2020{\natexlab{b}}.

\bibitem[Poon and Domingos(2011)]{poon2011sum}
Hoifung Poon and Pedro Domingos.
\newblock Sum-product networks: A new deep architecture.
\newblock In \emph{2011 IEEE International Conference on Computer Vision
  Workshops (ICCV Workshops)}, 2011.

\bibitem[Rahman et~al.(2014)Rahman, Kothalkar, and Gogate]{rahman2014cutset}
Tahrima Rahman, Prasanna Kothalkar, and Vibhav Gogate.
\newblock Cutset networks: A simple, tractable, and scalable approach for
  improving the accuracy of chow-liu trees.
\newblock In \emph{Joint European conference on machine learning and knowledge
  discovery in databases}, 2014.

\bibitem[Rooshenas and Lowd(2014)]{rooshenas2014learning}
Amirmohammad Rooshenas and Daniel Lowd.
\newblock Learning sum-product networks with direct and indirect variable
  interactions.
\newblock In \emph{Proceedings of the 31st International Conference on Machine
  Learning (ICML)}, 2014.

\bibitem[Ruan et~al.(2021)Ruan, Ullrich, Severo, Townsend, Khisti, Doucet,
  Makhzani, and Maddison]{ruan2021improving}
Yangjun Ruan, Karen Ullrich, Daniel~S Severo, James Townsend, Ashish Khisti,
  Arnaud Doucet, Alireza Makhzani, and Chris Maddison.
\newblock Improving lossless compression rates via monte carlo bits-back
  coding.
\newblock In \emph{Proceedings of the 38th International Conference on Machine
  Learning (ICML)}, 2021.

\bibitem[Shao et~al.(2022)Shao, Molina, Vergari, Stelzner, Peharz, Liebig, and
  Kersting]{shao2022conditional}
Xiaoting Shao, Alejandro Molina, Antonio Vergari, Karl Stelzner, Robert Peharz,
  Thomas Liebig, and Kristian Kersting.
\newblock Conditional sum-product networks: Modular probabilistic circuits via
  gate functions.
\newblock \emph{International Journal of Approximate Reasoning}, 2022.

\bibitem[Shih et~al.(2021)Shih, Sadigh, and Ermon]{shih2021hyperspns}
Andy Shih, Dorsa Sadigh, and Stefano Ermon.
\newblock Hyperspns: Compact and expressive probabilistic circuits.
\newblock In \emph{Advances in Neural Information Processing Systems 34
  (NeurIPS)}, 2021.

\bibitem[Townsend et~al.(2018)Townsend, Bird, and
  Barber]{townsend2018practical}
James Townsend, Thomas Bird, and David Barber.
\newblock Practical lossless compression with latent variables using bits back
  coding.
\newblock In \emph{Proceedings of the 6th International Conference on Learning
  Representations (ICLR)}, 2018.

\bibitem[Tran et~al.(2019)Tran, Vafa, Agrawal, Dinh, and
  Poole]{tran2019discrete}
Dustin Tran, Keyon Vafa, Kumar Agrawal, Laurent Dinh, and Ben Poole.
\newblock Discrete flows: Invertible generative models of discrete data.
\newblock In \emph{Advances in Neural Information Processing Systems 32
  (NeurIPS)}, 2019.

\bibitem[Ventola et~al.(2020)Ventola, Stelzner, Molina, and
  Kersting]{ventola2020residual}
Fabrizio Ventola, Karl Stelzner, Alejandro Molina, and Kristian Kersting.
\newblock Residual sum-product networks.
\newblock In \emph{Proceedings of the 10th International Conference on
  Probabilistic Graphical Models (PGM)}, 2020.

\bibitem[Vergari et~al.(2020)Vergari, Choi, Peharz, and Van~den
  Broeck]{PCTuto20}
Antonio Vergari, YooJung Choi, Robert Peharz, and Guy Van~den Broeck.
\newblock Probabilistic circuits: Representations, inference, learning and
  applications.
\newblock \emph{AAAI Tutorial}, 2020.

\bibitem[Vergari et~al.(2021)Vergari, Choi, Liu, Teso, and Van~den
  Broeck]{vergari2021compositional}
Antonio Vergari, YooJung Choi, Anji Liu, Stefano Teso, and Guy Van~den Broeck.
\newblock A compositional atlas of tractable circuit operations for
  probabilistic inference.
\newblock In \emph{Advances in Neural Information Processing Systems 34
  (NeurIPS)}, 2021.

\bibitem[Xiao et~al.(2017)Xiao, Rasul, and Vollgraf]{xiao2017fashionmnist}
Han Xiao, Kashif Rasul, and Roland Vollgraf.
\newblock Fashion-mnist: a novel image dataset for benchmarking machine
  learning algorithms.
\newblock \emph{arXiv preprint arXiv:1708.07747}, 2017.

\bibitem[Zhang et~al.(2021)Zhang, Juba, and Van~den Broeck]{ZhangICML21}
Honghua Zhang, Brendan Juba, and Guy Van~den Broeck.
\newblock Probabilistic generating circuits.
\newblock In \emph{Proceedings of the 38th International Conference on Machine
  Learning (ICML)}, 2021.

\bibitem[Ziegler and Rush(2019)]{ziegler2019latent}
Zachary Ziegler and Alexander Rush.
\newblock Latent normalizing flows for discrete sequences.
\newblock In \emph{Proceedings of the 36th International Conference on Machine
  Learning (ICML)}, 2019.

\end{thebibliography}

\newpage

\section*{Checklist}

\begin{enumerate}

\item For all authors...
\begin{enumerate}
  \item Do the main claims made in the abstract and introduction accurately reflect the paper's contributions and scope?
    \answerYes{} Contributions are clearly stated in the third and fourth paragraphs of the introduction.
  \item Did you describe the limitations of your work?
    \answerYes{} The proposed learning algorithms are designed for smooth and decomposable PCs.
  \item Did you discuss any potential negative societal impacts of your work?
    \answerNA{}
  \item Have you read the ethics review guidelines and ensured that your paper conforms to them?
    \answerYes{}
\end{enumerate}

\item If you are including theoretical results...
\begin{enumerate}
  \item Did you state the full set of assumptions of all theoretical results?
    \answerYes{} All theorems in the main text formally state all assumptions.
        \item Did you include complete proofs of all theoretical results?
    \answerYes{} Proofs are included in the appendix.
\end{enumerate}

\item If you ran experiments...
\begin{enumerate}
  \item Did you include the code, data, and instructions needed to reproduce the main experimental results (either in the supplemental material or as a URL)?
    \answerYes{} Experiment setup is detailed in the main text and the appendix.
  \item Did you specify all the training details (e.g., data splits, hyperparameters, how they were chosen)?
    \answerYes{} Training details are provided in the appendix.
        \item Did you report error bars (e.g., with respect to the random seed after running experiments multiple times)?
    \answerNA{}
        \item Did you include the total amount of compute and the type of resources used (e.g., type of GPUs, internal cluster, or cloud provider)?
    \answerYes{} Details about computational resources can be found in the appendix.
\end{enumerate}

\item If you are using existing assets (e.g., code, data, models) or curating/releasing new assets...
\begin{enumerate}
  \item If your work uses existing assets, did you cite the creators?
    \answerYes{} We cited all PC learning algorithms as well as the adopted datasets.
  \item Did you mention the license of the assets?
    \answerYes{} We specified that the used algorithms are publicly available.
  \item Did you include any new assets either in the supplemental material or as a URL?
    \answerYes{} We included our code in the supplementary material.
  \item Did you discuss whether and how consent was obtained from people whose data you're using/curating?
    \answerNA{}
  \item Did you discuss whether the data you are using/curating contains personally identifiable information or offensive content?
    \answerNA{}
\end{enumerate}

\item If you used crowdsourcing or conducted research with human subjects...
\begin{enumerate}
  \item Did you include the full text of instructions given to participants and screenshots, if applicable?
    \answerNA{}
  \item Did you describe any potential participant risks, with links to Institutional Review Board (IRB) approvals, if applicable?
    \answerNA{}
  \item Did you include the estimated hourly wage paid to participants and the total amount spent on participant compensation?
    \answerNA{}
\end{enumerate}

\end{enumerate}

\newpage
\appendix

\section{Pseudocode}
\label{app-sec:code}
In this section, we list the detailed algorithms for pruning  (Section~\ref{sec:prune}), growing  (Section~\ref{sec:struct-learn}), circuit flows computation (Definition~\ref{def:flow}), and mini-batch Expectation Maximization (Section~\ref{sec:struct-learn}).

Algorithm~\ref{alg:prune} shows how to prune $k$ percentage edges from PC $\PC$ following heuristic $h$.
\begin{algorithm}[th]
\caption{\label{alg:prune}\prune(\PC, $h$, $k$)}
\SetKwInOut{Input}{Input}
\SetKwInOut{Output}{Output}
\ResetInOut{Output}
\Input{a non-deterministic PC $\PC$, heuristic $h$ deciding which edge to prune, $h$ can be $\eflow$, $\erand$, or $\eparam$, 
percentage of edges to prune $k$
}
\Output{a PC \PC' after pruned}
$ \mathtt{old2new} \leftarrow $ mapping from input PC $n \in \PC$ to pruned PC\\
$s(n,c) \leftarrow$ compute a score for each edge $(n,c)$ based on heuristic $h$\\
$f(n,c) \leftarrow false$\\
$f(n,c) \leftarrow true$ if $s(n,c)$ ranks the last $k$\\
\tcp{visit children before parents}
\ForEach{$n \in \PC$}{
    \uIf{$n$ is a leaf}{
        $\mathtt{old2new}[n] \leftarrow n$
    }\uElseIf{$n$ is a sum}{
        $\mathtt{old2new}[n] \leftarrow \bigoplus
        ([\mathtt{old2new}(c) \text{ for } c \in \ch(n) \text{ and if } f(n, c)])$
    }\Else($n$ is a product){
        $\mathtt{old2new}[n] \leftarrow \bigotimes([\mathtt{old2new}(c) \text{ for } c \in \ch(n)])$
    }
}
\Return $\mathtt{old2new}[n_r]$ where $n_r$ is the root of \PC
\end{algorithm}

Algorithm~\ref{alg:grow} shows show a feedforward implementation of growing operation.

    \begin{algorithm}[H]
\SetAlgoLined
\caption{\grow(\PC, $\sigma^2$)}\label{alg:grow}
\SetKwInOut{Input}{Input}
\SetKwInOut{Output}{Output}
\ResetInOut{Output}
\Input{a PC $\PC$, Gaussian noisy variance $\sigma^2$}
\Output{a PC \PC' after growing operation}
$\mathtt{old2new} \leftarrow $ a dictionary mapping input PC units $n \in \PC$ to units of the growed PC\\
\ForEach($\;$ \texttt{// visit children before parents}){$n \in \PC$}{
    \textbf{if} $n$ is an input unit \textbf{then} $\mathtt{old2new}[n] \leftarrow (n, \mathsf{deepcopy}(n))$\\
    \uElse{$\mathtt{chs\_1}, \mathtt{chs\_2} \leftarrow [\mathtt{old2new}[c][0] \mathrm{~for~} c \mathrm{~in~} \ch(n)], \, [\mathtt{old2new}[c][1] \mathrm{~for~} c \mathrm{~in~} \ch(n)]$\\
        \textbf{if} $n$ is a product unit \textbf{then} $\mathtt{old2new}[n] \leftarrow (\bigotimes(\mathtt{chs\_1}), \bigotimes(\mathtt{chs\_2}))$\\
        \uElseIf{$n$ is a sum unit}{
            $n_1, n_2 \leftarrow \bigoplus([\mathtt{chs\_1},\mathtt{chs\_2}]), \bigoplus([\mathtt{chs\_1},\mathtt{chs\_2}])$\\
            $\params_{\given n_i} \leftarrow \mathsf{normalize}([\params_{\given n},\params_{\given n}]) \times \boldsymbol{\epsilon}) \quad \epsilon \sim \calN(\mathbf{1}, \mathbf{\sigma^2}) \mathrm{~for~} i \mathrm{~in~} [1,2]$\\
            $\mathtt{old2new}[n] \leftarrow (n_1, n_2)$
        }
    }
}
\textbf{return} $\mathtt{old2new}[r][0]\;$ \texttt{// $r$ is the root unit of $\PC$}
\end{algorithm}

\newpage
Algorithm~\ref{alg:exp-flow} computes the circuit flows of a sample $\x$ given PC $\PC$ with parameters $\params$ though one forward pass (line 1) and one backward pass (line 2-8).
\begin{algorithm}[!h]
 \caption{\label{alg:exp-flow}\textsf{CircuitFlow}($\PC$,$\params$,$\x$)}
\SetAlgoLined
    \SetKwInOut{Input}{Input}
    \SetKwInOut{Output}{Output}
    \Input{a PC $\PC$ with parameters $\params$; sample $\x$}
    \Output{circuit flow $\mathsf{flow}[n,c]$ for each edge $(n,c)$ and $\mathsf{flow}[n]$ for each node $n$}
    $\forall n \in \PC,$ $\mathsf{p}[n] \leftarrow \p_{n}(\x)$  computed as in Equation~\ref{eq:EVI}\\
    For root $n_r, \mathsf{flow}[n]\leftarrow 1$\\
    \For{$n\in \PC$ \text{ in backward order}}{
    $\mathsf{flow}[n] \leftarrow \sum_{g\in \pa(n)} \mathsf{flow}[g]$\\
    \uIf{n is a sum node}{
        $\forall c\in \ch(n), \mathsf{flow}[n,c] \leftarrow \theta_{c|n} \frac{\mathsf{p}[c]}{\mathsf{p}[n]}\mathsf{flow}[n] $
    }
    \Else{
    $\forall c\in \ch(n), \mathsf{flow}[n,c] \leftarrow \mathsf{flow}[n]$}
    }
\end{algorithm}

Algorithm~\ref{alg:mini-batchEM} shows the pipeline of mini-batches Expectation Maximization algorithm given PC $\PC$, dataset $\data$, batch size $B$ and learning rate $\alpha$.
\begin{algorithm}[!h]
 \caption{\label{alg:mini-batchEM}\textsf{StochasticEM}($\PC$,$\data$;$B$,$\alpha$)}
\SetAlgoLined
    \SetKwInOut{Input}{Input}
    \SetKwInOut{Output}{Output}
    \Input{a PC $\PC$; dataset $\data$; batch size $B$; learning rate $\alpha$}
    \Output{parameters $\params$ estimated from $\data$}
    $\theta \leftarrow $ random initialization\\
    For root $n_r, \mathsf{flow}[n]\leftarrow 1$\\
    \While {not converged or early stopped}{
        $\data' \leftarrow $ $B$ random samples from $\data$\\
        $\mathsf{flow} \leftarrow \sum_{\x \in \data'}\mathsf{CircuitFlow}(\PC, \params, \x)$\\
        \For{sum unit $n$ and its child $c$}{
        $\theta_{c|n}^{new} \leftarrow \mathsf{flow}[n,c]/\mathsf{flow}[n] $\\
        $\theta_{c|n} \leftarrow \alpha\theta_{c|n}^{(new)} + (1-
\alpha)\theta_{c|n}$
        }
    }
\end{algorithm}

\section{Proofs}
\label{app-sec:proof}

In this section, we provide detailed proofs of Theorem~\ref{thm:prune_one_ll} (Section~\ref{app-sec:proof-single}) and Theorem~\ref{thm:prune-multi} (Section~\ref{app-sec:proof-multi}).

\subsection{Pruning One Edge over One Example}
\label{app-sec:proof-single}

\begin{lem}[Pruning One Edge Log-Likelihood Lower Bound]
\label{lem:one_edge_ll}
For a PC $\PC$ and a sample $\x$, the loss of log-likelihood by pruning away edge $(n,c)$ is
    \begin{align*}
        \Delta\LL(\{\x\},\PC,\{(n,c)\})
        =  \log\left(\frac{1-\theta_{c\given n}}{1 \!-\! \theta_{c\given n} \!+\! \theta_{c\given n} \flow_n(\x) \!-\! \flow_{n,c}(\x)}\right) 
        \leq - \log(1 \!-\! \flow_{n,c}(\x)).
    \end{align*}
\end{lem}

\begin{proof}
For notation simplicit, denote the probability of units $m$ (resp. $n$) in the original (resp. pruned) PC given sample $\x$ as $\p_{m}(\x)$ (resp. $\p'_{n}(\x)$). As a slight extension of Definition~\ref{def:flow}, we define $F_{n}(\x; m)$ as the flow of unit $n$ \wrt the PC rooted at $m$.

The proof proceeds by induction over the PC's root unit. That is, we first consider pruning $(n,c)$ \wrt the PC rooted at $n$. Then, in the induction step, we prove that if the lemma holds for PC rooted at $m$, then it also holds for PC rooted at any parent unit of $m$. Instead of directly proving the statement in Lemma~\ref{lem:one_edge_ll}, we first prove that for any root node $m$, the following holds:
    \begin{align*}
        \p_{m} (\x) - \p'_{m} (\x) = F_{n} (\x; m) \cdot p_m(\x) \cdot \bigg ( \frac{1}{1-\theta}\frac{\flow_{n,c}(\x; m)}{\flow_{n}(\x; m)} - \frac{\theta}{1-\theta} \bigg ). \numberthis\label{eq:lem1-proof-eq0}
    \end{align*}

Base case: pruning an edge of the root unit. That is, the root unit of the PC is $n$. In this case, we have
    \begin{align*}
        \p_n(\x) - \p'_n(\x) &=  \sum_{c' \in \ch(n)} \theta_{c'\given n} \cdot p_c(\x) - \sum_{c' \in \ch(n)\backslash c}\theta_{c'\given n}' \cdot p_c'(\x) \\
        &= \theta_{c\given n} \cdot p_c(\x) + \sum_{c' \in \ch(n)\backslash c} \theta_{c'\given n} \cdot p_c(\x) - \sum_{c' \in \ch(n)\backslash c} \theta_{c'\given n}' \cdot p_c(\x),\numberthis\label{eq:lem1-proof-eq1}
    \end{align*}
\noindent where $\theta_{c\given n}'$ denotes the normalized parameter corresponding to edge $(n, c)$ in the pruned PC.
Specifically, we have
    \begin{align*}
        \forall m \in \ch(n)\backslash c, \quad \theta'_{m\given n}=\frac{\theta_{m\given n}}{\sum_{c'\in \ch(n)\backslash c}\theta_{c'\given n}} = \frac{\theta_{m\given n}}{1-\theta_{c\given n}}.
    \end{align*}
For notation simplicity, denote $\theta := \theta_{c\given n}$. Plug in the above definition into Equation~\ref{eq:lem1-proof-eq1}, we have
\begin{align*}
    \p_n (\x) - \p'_{n} (\x) &= \theta_{c\given n} \cdot p_c(\x) +\sum_{c' \in \ch(n)\backslash c} \theta_{c'\given n} \cdot p_c(\x) - \frac{1}{1-\theta}\sum_{c' \in \ch(n)\backslash c}\theta_{c'\given n} \cdot p_c(\x) \\
    & = \theta_{c\given n} \cdot p_c(\x) - \frac{\theta}{1-\theta}\sum_{c' \in \ch(n)\backslash c}\theta_{c'\given n} \cdot p_c(\x) \\
    & = \theta_{c\given n} \cdot p_c(\x) - \frac{\theta}{1-\theta}(p_n(\x) - \theta_{c\given n} p_c(\x)) \\
    & = \frac{1}{1-\theta} \cdot \theta_{c\given n} \cdot p_c(\x) - \frac{\theta}{1-\theta} \cdot p_n (\x) \\
    & \overset{(a)}{=} \frac{1}{1-\theta}  \cdot p_n(\x) \cdot \frac{\flow_{n,c}(\x; n)}{\flow_{n}(\x; n)} - \frac{\theta}{1-\theta} \cdot p_n(\x) \\
    & = F_n (\x; n) \cdot p_n(\x) \cdot \bigg ( \frac{1}{1-\theta}\frac{\flow_{n,c}(\x; n)}{\flow_{n}(\x; n)} - \frac{\theta}{1-\theta} \bigg ), \numberthis\label{eq:lem1-proof-eq2} \\
\end{align*}
\noindent where $(a)$ follows from the fact that $F_n(\x; n) = 1$ and $F_{n,c} (\x; n) = \theta_{c\given n} \p_{c} (\x) / \p_{n} (\x)$.

Inductive case \#1: suppose Equation~\ref{eq:lem1-proof-eq0} holds for $m$. If product unit $d$ is a parent of $m$, we show that Equation~\ref{eq:lem1-proof-eq0} also holds for $d$:

\begin{align*}
    p_{d}(\x) - p'_{d}(\x) & = \prod_{n'\in \ch(d)}p_{n'}(\x) - \prod_{n'\in \ch(d)}p'_{n'}(\x) \\
    & = (p_m(\x) - p'_m(\x)) \prod_{n'\in \ch(d)\backslash m}p_{n'}(\x) \\
    & \overset{(a)}{=} F_n (\x; m) \cdot p_m(\x) \cdot \bigg ( \frac{1}{1-\theta}\frac{\flow_{n,c}(\x; m)}{\flow_{n}(\x; m)} - \frac{\theta}{1-\theta} \bigg ) \cdot \prod_{n'\in \ch(d)\backslash m}p_{n'}(\x) \\
    & \overset{(b)}{=} F_n (\x; d) \cdot p_{d}(\x) \cdot \bigg ( \frac{1}{1-\theta}\frac{\flow_{n,c}(\x;d)}{\flow_{n}(\x;d)} - \frac{\theta}{1-\theta} \bigg ),
\end{align*}
\noindent where $(a)$ is the inductive step that applies Equation~\ref{eq:lem1-proof-eq2}; $(b)$ follows from the fact that (note that $d$ is a product unit) $F_{n}(\x; m) = F_{n}(\x; d)$ and $F_{n,c}(\x; m) = F_{n,c}(\x; d)$.

Inductive case \#2: for sum unit $d$, suppose Equation~\ref{eq:lem1-proof-eq0} holds for $m$, where $m \in \calA$ iff $m \in \ch(d)$ and $m$ is an ancester of $n$ and $c$. Assume all other children of $d$ are not ancestoer of $n$, we show that Equation~\ref{eq:lem1-proof-eq0} also holds for $d$:
\begin{align*}
    p_d(\x) - p'_d(\x) & = \theta_{m\given d} \cdot (p_{m}(\x)-p'_{m}(\x))\\
    & = \theta_{m\given d} \cdot F_n (\x; m) \cdot p_{m}(\x) \cdot \bigg ( \frac{1}{1-\theta}\frac{\flow_{n,c}(\x; m)}{\flow_{n}(\x; m)} - \frac{\theta}{1-\theta} \bigg ) \\
    & = \theta_{m\given d} \cdot F_n (\x; m) \cdot p_{m}(\x) \cdot \bigg ( \frac{1}{1-\theta}\frac{\flow_{n,c}(\x; d)}{\flow_{n}(\x; d)} - \frac{\theta}{1-\theta} \bigg ) \\
    & = \theta_{m\given d} \cdot F_n (\x; d) \cdot \frac{\sum_{m' \in \ch(d)} \theta_{m'\given d} \p_{m'} (\x)}{\theta_{m\given d} \p_m (\x)} \cdot p_{m}(\x) \cdot \bigg ( \frac{1}{1-\theta}\frac{\flow_{n,c}(\x; d)}{\flow_{n}(\x; d)} - \frac{\theta}{1-\theta} \bigg ) \\
    & = F_n (\x; d)  \cdot \bigg ( \sum_{m' \in \ch(d)} \theta_{m'\given d} \p_{m'} (\x) \bigg ) \cdot \bigg ( \frac{1}{1-\theta}\frac{\flow_{n,c}(\x; d)}{\flow_{n}(\x; d)} - \frac{\theta}{1-\theta} \bigg ) \\
    & = F_n (\x; d) \cdot \p_{d} (\x) \cdot \bigg ( \frac{1}{1-\theta}\frac{\flow_{n,c}(\x; d)}{\flow_{n}(\x; d)} - \frac{\theta}{1-\theta} \bigg ).
\end{align*}

Therefore, following Equation~\ref{eq:lem1-proof-eq0} for root $r$, we have
\begin{align*}
    & \frac{\p_{r} (\x) - \p'_{r} (\x)}{p_r(\x)} =  \frac{1}{1-\theta} \flow_{n,c}(\x; r) - \frac{\theta}{1-\theta} F_{n} (\x; r) \\
    \Leftrightarrow \quad & \frac{\p'_{r} (\x)}{p_r(\x)} = 1 + \frac{\theta}{1-\theta} F_{n} (\x; r) - \frac{1}{1-\theta} \flow_{n,c}(\x; r)
\end{align*}
Therefore, we have 
\begin{align*}
    \Delta\LL(\{\x\},\PC,\{(n,c)\}) & = \log \p_r (\x) - \log \p'_r (\x) \\
    & = \frac{1}{|\data|}\sum_{\x\in\data} \log\left(\frac{1-\theta_{c\given n}}{1 \!-\! \theta_{c\given n} \!+\! \theta_{c\given n} \flow_n(\x; r) \!-\! \flow_{n,c}(\x; r)}\right) \\
    & \overset{(a)}{\leq} - \log(1 - F_{n,c}(\x)),
\end{align*}
\noindent where $(a)$ follows from the fact that $F_{n,c}(\x) \leq F_{n}(\x)$.
\end{proof}

Theorem~\ref{thm:prune_one_ll} follows directly from Lemma~\ref{lem:one_edge_ll} by noting that for any dataset $\data$, $\Delta\LL(\data,\PC,\{(n,c)\}) = \frac{1}{\abs{\data}} \Delta\LL(\{\x\},\PC,\{(n,c)\})$.

\subsection{Pruning Multiple Edges}
\label{app-sec:proof-multi}

\begin{proof}
Similar to the proof of Lemma~\ref{lem:one_edge_ll}, we prove Theorem~\ref{thm:prune-multi} by induction. Different from Lemma~\ref{lem:one_edge_ll}, we induce a slightly different objective:
    \begin{align*}
        \p_{m} (\x) - \p'_{m} (\x) \leq \sum_{(n,c)\in \edges \cap \des(m)} F_{n} (\x; m) \cdot p_m(\x) \cdot \bigg ( \frac{1}{1-\theta_{c\given n}}\frac{\flow_{n,c}(\x; m)}{\flow_{n}(\x; m)} - \frac{\theta_{c\given n}}{1-\theta_{c\given n}} \bigg ),\numberthis\label{eq:thm2-proof-eq1}
    \end{align*}
\noindent where $\des(n)$ is the set of descendent units of $n$.

Base case: the base case follows directly from the proof of Lemma~\ref{lem:one_edge_ll}, and lead to the conclusion in Equation~\ref{eq:lem1-proof-eq2}.

Inductive case \#1: suppose for all children of a product unit $d$, Equation~\ref{eq:thm2-proof-eq1} holds, we show that Equation~\ref{eq:thm2-proof-eq1} also holds for $d$:
    \begin{align*}
        \p_d (\x) - \p'_d (\x) & = \prod_{m \in \ch(d)} \p_m (\x) - \prod_{m \in \ch(d)} \p'_m (\x) \\
        & = \prod_{m \in \ch(d)} \p_m (\x) - \prod_{m \in \ch(d)} \Big ( \p_m (\x) - ( \p_m (\x) - \p'_m (\x) ) \Big ) \\
        & \leq \sum_{m \in \ch(d)} \Big ( \p_m (\x) - \p'_m (\x) ) \Big ) \cdot \prod_{m' \in \ch(d) \backslash m} \p_{m'} (\x) \\
        & \overset{(a)}{\leq} \sum_{m \in \ch(d)} \sum_{(n,c)\in \edges \cap \des(m)} F_{n} (\x; d) \cdot p_d(\x) \cdot \bigg ( \frac{1}{1-\theta_{c\given n}}\frac{\flow_{n,c}(\x; m)}{\flow_{n}(\x; m)} - \frac{\theta_{c\given n}}{1-\theta_{c\given n}} \bigg ) \\
        & \leq \sum_{(n,c)\in \edges \cap \des(d)} F_{n} (\x; d) \cdot p_d(\x) \cdot \bigg ( \frac{1}{1-\theta_{c\given n}}\frac{\flow_{n,c}(\x; d)}{\flow_{n}(\x; d)} - \frac{\theta_{c\given n}}{1-\theta_{c\given n}} \bigg ),
    \end{align*}
\noindent where $(a)$ uses the definition that $\p_{d} (\x) = \prod_{m \in \ch(d)} \p_{m} (\x)$.

Inductive case \#2: suppose for all children of a sum unit $d$, Equation~\ref{eq:thm2-proof-eq1} holds, we show that Equation~\ref{eq:thm2-proof-eq1} also holds for $d$:
    \begin{align*}
        \p_{d} (\x) - \p'_{d} (\x) & = \sum_{m \in \ch(d) \cap (d,m) \not\in \edges} \theta_{m\given d} \cdot \Big ( \p_{m} (\x) - \p'_{m} (\x) \Big ) + \sum_{m \in \ch(d) \cap (d,m) \in \edges} \theta_{m\given d} \cdot \Big ( \p_{m} (\x) - \p'_{m} (\x) \Big ) \\
        & \overset{(a)}{=} \sum_{m \in \ch(d) \cap (d,m) \not\in \edges} \theta_{m\given d} \cdot \Big ( \p_{m} (\x) - \p'_{m} (\x) \Big ) \\
        & \qquad + \sum_{m \in \ch(d) \cap (d,m) \in \edges} \theta_{m\given d} \cdot F_{n} (\x; m) \cdot p_m(\x) \cdot \bigg ( \frac{1}{1-\theta_{c\given n}}\frac{\flow_{n,c}(\x; m)}{\flow_{n}(\x; m)} - \frac{\theta_{c\given n}}{1-\theta_{c\given n}} \bigg ),
    \end{align*}
\noindent where $(a)$ follows from the base case of the induction. Next, we focus on the first term of the above equation:
    \begin{align*}
        & \sum_{m \in \ch(d) \cap (d,m) \not\in \edges} \theta_{m\given d} \cdot \Big ( \p_{m} (\x) - \p'_{m} (\x) \Big ) \\
        \leq & \sum_{m \in \ch(d) \cap (d,m) \not\in \edges} \sum_{(n,c) \in \edges \cap \des(m)} \theta_{m\given d} \cdot \Big ( \p_{m} (\x) - \p'_{m} (\x) \Big ) \\
        \leq & \sum_{m \in \ch(d) \cap (d,m) \not\in \edges} \sum_{(n,c) \in \edges \cap \des(m)} \theta_{m\given d} \cdot F_{n} (\x; m) \cdot p_m(\x) \cdot \bigg ( \frac{1}{1-\theta_{c\given n}}\frac{\flow_{n,c}(\x; m)}{\flow_{n}(\x; m)} - \frac{\theta_{c\given n}}{1-\theta_{c\given n}} \bigg ) \\
        \leq & \sum_{(n,c) \in \edges \cap \des(d)} F_{n} (\x; d) \cdot p_d(\x) \cdot \bigg ( \frac{1}{1-\theta_{c\given n}}\frac{\flow_{n,c}(\x; d)}{\flow_{n}(\x; d)} - \frac{\theta_{c\given n}}{1-\theta_{c\given n}} \bigg ),
    \end{align*}
\noindent where the derivation of the last inequality follows from the corresponding steps in the proof of Lemma~\ref{lem:one_edge_ll}.

Therefore, from Equation~\ref{eq:thm2-proof-eq1}, we can conclude that 
    \begin{align*}
        \Delta\LL(\data,\PC,\edges)
        \leq - \frac{1}{\abs{\data}}\sum_{\x}\log(1-\sum_{(n, c) \in \calE} \flow_{n,c}(\x)).
    \end{align*}

Finally, we prove the approximation step in Equation~\ref{eq:prune_mul_d_ll}. Let $\epsilon(\cdot)=\sum_{(n, c) \in \calE} \flow_{n,c}(\cdot)\in [0,1)$. We have,
\begin{align*}
\text{RHS}&
=-\sum_{\x \in \data}\log(1-\epsilon(\x)) 
= -\sum_{\x \in \data}\sum_{k=1}^{\infty}-\frac{\epsilon(\x)^k}{k}(\text{Taylor expansion})
\leq\sum_{\x \in \data}\sum_{k=1}^{\infty}\epsilon(\x)^k\\
&=\sum_{\x \in \data}\frac{\epsilon(\x)}{1-\epsilon(\x)}
=\frac{1}{1-\epsilon}\sum_{\x \in \data}\epsilon(\x)
=\frac{1}{1-\epsilon}\sum_{(n, c) \in \calE} \sum_{\x \in \data}\flow_{n,c}(\x)
=\frac{1}{1-\epsilon}\sum_{(n, c) \in \calE} \flow_{n,c}(\data).
\end{align*}
\end{proof}

\section{Experiments Details}
\label{app-sec:exp}
\paragraph{Hardware specifications} All experiments are performed on a server with 32 CPUs, 126G Memory, and NVIDIA RTX A5000 GPUs with 26G Memory. In all experiments, we only use a single GPU on the server.
\subsection{Datasets}
For MNIST-family datasets, we split 5\% of training set as validation set for early stopping. For Penn Tree Bank dataset, we follow the setting in~\citet{mikolov2012subword} to split a training, validation, and test set. Table~\ref{tab:data} lists the all the dataset statistics.
\begin{table}[h]
    \centering
    \caption{Dataset statistics including number of variables~(\textbf{\#vars}), number of categories for each variable~(\textbf{\#cat}), and number of samples for training, validation and test set~(\textbf{\#train}, \textbf{\#valid}, \textbf{\#test}).}
    \begin{tabular}{l|rrrrr}\toprule
        \textbf{Dataset} & $n$~(\textbf{\#vars}) & $k$~(\textbf{\#cat}) & \textbf{\#train} & \textbf{\#valid} & \textbf{\#test}\\\midrule
        MNIST & 28$\times$28 & 256 & 57000 & 3000 & 10000\\
        EMNIST(MNIST) & 28$\times$28 & 256  & 57000 & 3000 & 10000\\
        EMNIST(Letters) & 28$\times$28 & 256 & 118560 & 6240 & 20800\\
        EMNIST(Balanced) & 28$\times$28 & 256 & 107160 & 5640 & 18800\\
        EMNIST(ByClass) & 28$\times$28 & 256 & 663035 & 34897 & 116323\\
        FashionMNIST & 28$\times$28 & 256 & 57000 & 3000 & 10000\\\midrule
        Penn Tree Bank & 288 & 50 & 42068 & 3370 & 3761\\\bottomrule
    \end{tabular}
    \label{tab:data}
\end{table}
\subsection{Learning Hidden Chow-Liu Trees}
\label{app-sec:hclt}
\paragraph{HCLT structures.} Adopting hidden chow liu tree~(HCLT) PC architecture as in~\citet{LiuNeurIPS21}, we reimplement the learning process to speed it up and use a different training pipeline and hyper-parameters tuning.

\paragraph{EM parameter learning}
We adopt the EM parameter learning algorithm introduced in~\citet{ChoiAAAI21}, which computes the EM update target parameters using circuit flows. We use a stochastic mini-batches EM algorithm. Denoting $\theta^\new$ as the EM update target computed from a mini-batch of samples, and we update the targeting parameter with a learning rate $\alpha$: $\theta^{t+1} \leftarrow \alpha\theta^\new + (1-
\alpha)\theta^{t}$. $\alpha$ is piecewise-linearly annealed from $[1.0,0.1]$, $[0.1, 0.01]$, $[0.01, 0.001]$, and each piece is trained $T$ epochs.

\paragraph{Hyper-parameters searching.}For all the experiments, the hyper-parameters are searched from
\begin{itemize}
    \item $h\in \{8,16,32,64,128,256\}$, the hidden size of HCLT structures;
    \item $\gamma \in \{0.0001, 0.001, 0.01, 0.1, 1.0\}$, Laplace smoothing factor;
    \item $B \in \{128, 256, 512, 1024\}$, batch-size in mini-batches EM algorithm;
    \item $\alpha$ piecewise-linearly annealed from $[1.0,0.1]$, $[0.1, 0.01]$, $[0.01, 0.001]$, where each piece is called one mini-batch EM phase. Usually the algorithm will start to overfit as validation set and stop at the third phase;
    \item $T=100$, number of epochs for each mini-batch EM phase.
\end{itemize}
The PC size is quadratically growing with hidden size $h$, thus it is inefficient to do a grid search among the entire hyper-parameters space. What we do is to fist do a grid search when $h=8$ or $h=16$ to find the best Laplace smoothing factor $\gamma$ and batch-size $B$ for each dataset, and then fix $\gamma$ and $B$ to train a PC with larger hidden size $h\in\{32,64,128,256\}$. The best tuned $B$ is in $\{256,512\}$, which is different for different hidden size $h$, and the best tuned  $\gamma$ is $0.01$.
\subsection{Details of Section~\ref{sec:exp-benchmark}}
\label{app-sec:benchmark}
\paragraph{Sparse PC (ours).} Given an HCLT learned in Section~\ref{app-sec:hclt} as initial PC, we use the structure learning process proposed in Section~\ref{sec:struct-learn}. Specifically, starts from initial HCLT, for each iteration, we (1) prune 75\% of the PC parameters, and (2) grow PC size with Gaussian variance $\epsilon$, (3) finetuing PC using mini-batches EM parameter learning with learning rate $\alpha$. We prune and grow PC iteratively until the validation set likelihood is overfitted . The hyper-parameters are searched from 
\begin{itemize}
    \item $\epsilon \in \{0.1, 0.3, 0.5\}$, Gaussian variance in growing operation;
    \item $\alpha$, piecewise-linearly annealed from $[0.1, 0.01]$, $[0.01, 0.001]$;
    \item $T=50$, number of epochs for each mini-batch EM phase;
    \item for $\gamma$ and $B$, we use the tuned best number from Section~\ref{app-sec:hclt}.
\end{itemize}

\paragraph{HCLT.} The HLCT experiments in Table~\ref{tab:mnist} are performed following the original paper (Code \url{https://github.com/UCLA-StarAI/Tractable-PC-Regularization}), which is different from the leaning pipeline we use as our inital PC (Section~\ref{app-sec:hclt}).
\paragraph{SPN.} We reimplement the SPN architecture ourselves following~\citet{peharz2020random} and train it with the same mini-batch pipeline as HCLT.

\paragraph{IDF.}
We run all experiments with the code in the GitHub repo provided by the authors. We adopt an IDF model with the following hyperparameters: 8 flow layers per level; 2 levels; densenets with depth 6 and 512 channels; base learning rate 0.001; learning rate decay 0.999. The algorithm adopts an CPU-based entropy coder rANS.
\paragraph{BitSwap.}
We train all models using the following author-provided script: \url{https://github.com/fhkingma/bitswap/blob/master/model/mnist_train}.

\paragraph{BB-ANS.}
All experiments are performed using the following official code \url{https://github.com/bits-back/bits-back}.

\paragraph{McBits.}
All experiments are performed using the following official code \url{https://github.com/ryoungj/mcbits}.

\subsection{Details of Section~\ref{sec:exp-ablate}}
For all experiments in Section~\ref{sec:exp-ablate}, we use the best tuned $\gamma$ and $B$ from Section~\ref{app-sec:hclt} and hidden size $h$ ranging from $\{16,32,64,128\}$. For experiments ``What is the Smallest PC for the Same Likelihood?'', the hyper-parameters are searched from 
\begin{itemize}
    \item $k \in \{0.05, 0.1, 0.3\}$, percentage of parameters to prune each iteration;
    \item $\alpha$, piecewise-linearly annealed from $[0.3,0.1]$, $[0.1, 0.01]$, $[0.01, 0.001]$;
    \item $T=50$, number of epochs for each mini-batch EM phase;
\end{itemize}
For experiments ``What is the Best PC Given the Same Size?'', we use the same setting as in Section~\ref{app-sec:benchmark}.
\end{document}